\documentclass{article}
\usepackage[utf8]{inputenc}
\usepackage{authblk}
\usepackage{setspace}
\usepackage[margin=1in]{geometry}
\usepackage{graphicx}
\graphicspath{ {./figures/} }
\usepackage{subcaption}
\usepackage{lineno}
\usepackage[T1]{fontenc}    
\usepackage{hyperref}       
\usepackage{url}            
\usepackage{booktabs}       
\usepackage{amsfonts}       
\usepackage{nicefrac}       
\usepackage{microtype}      
\usepackage{lipsum}		
\usepackage{doi}
\usepackage{array}
\usepackage{amsfonts}
\usepackage{amsthm, amsmath}
\usepackage{amssymb}
\usepackage{mathtools}
\usepackage{latexsym}
\usepackage{inputenc}
\usepackage{graphicx}
\usepackage{multirow}
\usepackage{csquotes}
\usepackage{algorithm}
\usepackage{epsfig}
\usepackage{hyperref}
\usepackage{caption}
\usepackage{subcaption}
\usepackage{float}
\usepackage[noend]{algpseudocode}
\usepackage{color}
\usepackage[dvipsnames]{xcolor}
\usepackage{soul}

\newcommand{ \mb }[1]{ \mathbf{#1} }

\newcommand{ \bSigma }{ \boldsymbol{\Sigma} }

\newcommand{ \bLambda }{ \boldsymbol{\Lambda} }

\newcommand{ \bGamma }{ \boldsymbol{\Gamma} }
\newcommand{ \bPsi }{ \boldsymbol{\Psi} }
\newcommand{ \bpsi }{ \boldsymbol{\psi} }
\newcommand{ \bTheta }{ \boldsymbol{\Theta} }
\newcommand{ \bOmega }{ \boldsymbol{\Omega} }

\newcommand{ \verts }[1]{ \lvert #1 \rvert }

\newcommand{ \sm }{\setminus}

\usepackage[style=ieee, 
citestyle=numeric-comp,
sorting=none]{biblatex}

\title{Scalable Bigraphical Lasso: Two-way Sparse Network Inference for Count Data}

\author[1]{Sijia Li}
\author[1]{Mart\'in L\'opez-Garc\'ia}
\author[2]{Neil D. Lawrence}
\author[1*]{Luisa Cutillo}
\affil[1]{School of Mathematics, University of Leeds, Leeds, UK, LS2 9JT. }
\affil[2]{Department of Computer Science and Technology, University of Cambridge, Cambridge, UK, CB3 0FD.}
\affil[*]{Corresponding author. Email:  L.Cutillo@leeds.ac.uk}

\date{}

\begin{document}

\maketitle

\begin{abstract}
\begin{center}
\includegraphics[width=0.7\textwidth]{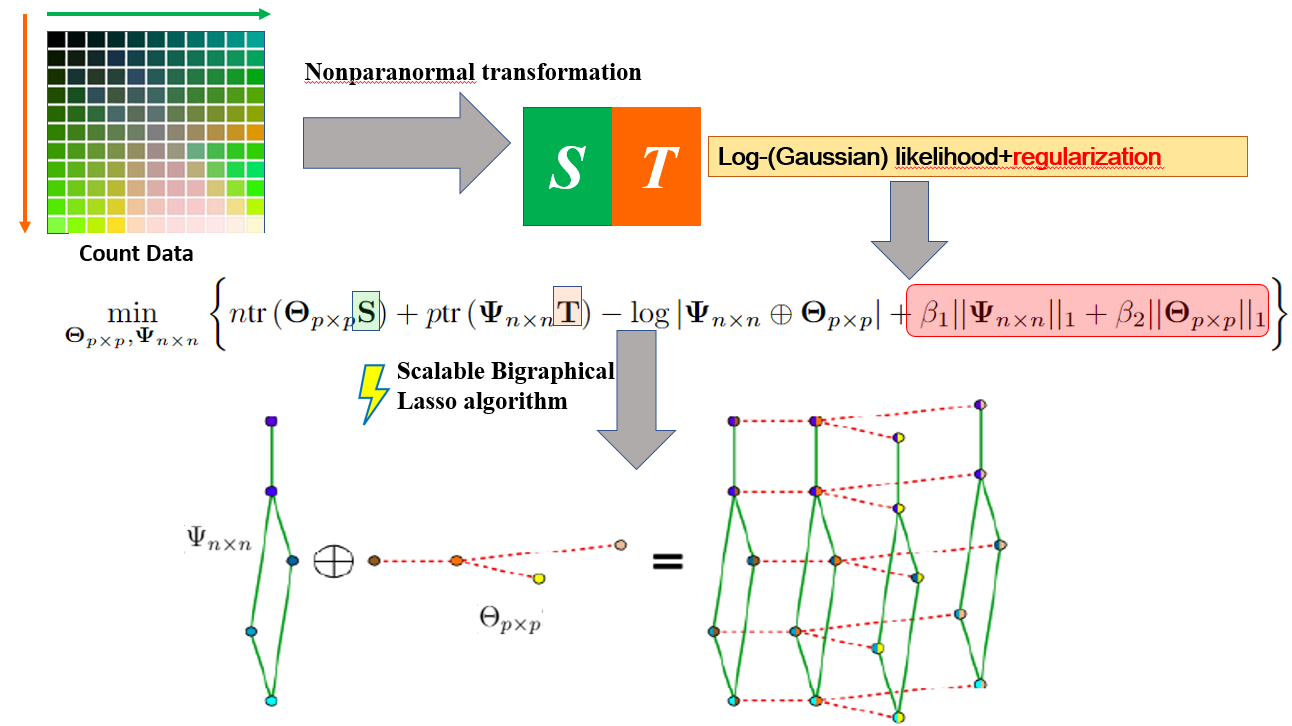}
\end{center}
\par Classically, statistical datasets  have a larger number of data points than features ($n > p$). The standard model of classical statistics caters for the case where data points are
considered conditionally independent given the parameters. However,
 for $n \approx p$ or $p>n$ such models are poorly determined. 
\cite{kalaitzis2013bigraphical} introduced the Bigraphical
 Lasso, an estimator for sparse precision matrices based on the
 Cartesian product of graphs. Unfortunately, the original Bigraphical Lasso algorithm is not applicable in case of large $p$ and $n$ due to memory
 requirements. We exploit eigenvalue decomposition of the Cartesian
 product graph to present a more efficient version of the algorithm
 which reduces memory requirements from $O(n^2p^2)$ to $O(n^2 + p^2)$. Many datasets in different application fields, such as biology, medicine and social science, come with count data, for which Gaussian based models are not applicable. Our multi-way network inference approach can be used for discrete data. 

\par Our methodology accounts for the dependencies across both instances and features, reduces the computational complexity for high dimensional data and enables to deal with both discrete and continuous data. Numerical studies on both synthetic and real datasets are presented to showcase the performance of our method.

\end{abstract}

\section{INTRODUCTION}
\par In this research, we develop a tensor-decomposition based two-way network inference approach for count data. Firstly, we present a Scalable Bigraphical Lasso algorithm, reducing both the space complexity and the computational complexity of the inference. Secondly, we extend the Bigraphical model to count data by means of a semiparametric approach. 
Our proposed methodology not only accounts for the dependencies across both instances and features, but also reduces the computational complexity for high dimensional data.
\par The main motivation of this research is that real world problems often come with correlations between several dimensions. Recently, Gaussian graphical models have been developed with tensor decomposition for multi-way network inference. 
For example, \cite{tsiligkaridis2013covariance} and \cite{zhou2014gemini} studied a matrix normal distribution where the precision matrix corresponds to the Kronecker product between the row-specific and the column-specific precision matrices. \cite{kalaitzis2013bigraphical} introduced Bigraphical Lasso, and \cite{greenewald2019tensor} introduced TeraLasso, both studying a multivariate normal distribution where the precision matrix corresponds to a Kronecker sum instead.

\par Many datasets in different application fields come with count data, for which Gaussian based models are not applicable. Some methods use other distributions to infer network from the data. \cite{jia2017learning} infers the gene regulation networks with a Poisson-Gamma based Bayesian Hierarchical Model, borrowing information across cells. \cite{mcdavid2019graphical} infers the gene regulation networks with a multivariate Hurdle model (zero-inflated mixed Gaussian). Several approaches have extended the use of Gaussian models to an appropriate continuous transformation of count data. \cite{liu2009nonparanormal} and \cite{liu2012high} proposed a semiparametric approach, and  \cite{roy2020nonparametric} proposed a nonparametric approach, while \cite{chiquet2019variational} considered Bayesian Hierarchical Models. However, all these methods only produce a one-way network inference. \cite{bartlett2021two} proposed a Bayesian model with a prior having decoupled two-way sparsity to infer a dynamic network structure through time, however, the method still depends on a pre-inferred or known ordering of time. Our method extends the Gaussian Copula transformation to enable a two-way network inference, where the structure in both dimensions is to be inferred simultaneously.
\par This paper is structured as follows: In Section 2 we present a detailed review on relevant literature; In Section 3 we present our Scalable Bigraphical Lasso algorithm for Gaussian data; In Section 4 we propose a semiparametric extension to the Bigraphical model for count data; In Section 5 we showcase the performance of our method on both synthetic and real datasets.


\section{BACKGROUND}
\label{sec:headings}
\subsection{From the matrix normal model to the Kronecker sum structure}
For a Gaussian density, a sparse precision matrix defines a weighted undirected graph in Gaussian
Markov random field relationship \cite{lauritzen1996graphical}, encoding conditional independence between variables in the Gaussian model. Therefore we can induce the network structure from the support of the precision matrix.
\par A matrix normal model with the Kronecker sum structure was proposed in \cite{kalaitzis2013bigraphical}. If a $p\times n$ random matrix $\bf{Y}$ follows a matrix normal distribution,
\begin{equation*}
    {\bf Y}\sim\mbox{\bf MN}_{p\times n}\left({\bf M};\bPsi_{n\times n}^{-1},\bTheta_{p\times p}^{-1}\right),
\end{equation*}
with ${\bf M}$ a $p\times n$ matrix, and with precision matrix $\bPsi_{n\times n}$ indicating the dependency structure in rows, and precision matrix $\bTheta_{p\times p}$ indicating the dependency structure in columns, the model can be reparametrized such that the vectorised random matrix follows a $np$-dimensional multivariate normal distribution (denoted as $\mbox{\bf mN}$):
\[\mbox{vec}\left(\bf{Y}\right)\sim\mbox{\bf{mN}}_{np}\left(\boldsymbol{0}_{np},\left(\bPsi_{n\times n}\otimes \bTheta_{p\times p}\right)^{-1}\right),\]
where $\otimes$ denotes the Kronecker product (\textit{KP}), $\bPsi_{n\times n}\otimes \bTheta_{p\times p}$ is the overall precision matrix, and $\boldsymbol{0}_{np}$ is a column vector of zeros of length $np$. \cite{kalaitzis2013bigraphical} proposed to use the Kronecker sum (\textit{KS}) $\bPsi_{n\times n}\oplus \bTheta_{p\times p}=\bPsi_{n\times n}\otimes I_{p}+I_n\otimes \bTheta_{p\times p}$ to structure the overall precision matrix. 
In a KS-structured matrix normal distribution, for a $p\times n$ random matrix $\bf{Y}$, we write
\[\mbox{vec}\left(\bf{Y}\right)\sim\mbox{\bf{mN}}_{np}\left(\boldsymbol{0}_{np},\left(\bPsi_{n\times n}\oplus \bTheta_{p\times p}\right)^{-1}\right).\]
The KS-structure has several advantages. Firstly, in algebraic graph theory, the Kronecker sum corresponds to the Cartesian product of graphs \cite{sabidussi1959graph}. A KS-structured model therefore provides intuitive and interpretable results. Secondly, for high-dimensional data, the KS-structure enhances the sparsity of the network, reducing the computation complexity and the memory requirement. 
\subsection{Rank-based estimation in a Gaussian graphical model}
To model count data or other non-Gaussian data in a Gaussian graphical model, the Gaussian copula can be applied to transfer these data into a latent Gaussian variable. \cite{liu2012high} proposed a semiparametric Gaussian copula for one-way network inference. For a $p\times n$ matrix $\bf{Y}$, \cite{liu2012high} considered it as $n$ samples of a $p-$dimensional vector $\left(Y_{1j},\dots,Y_{pj}\right)$. \cite{liu2012high} assumed that there exist functions $f=\left\{f_i\right\}_{i=1}^p$ such that for $j=1,\dots,n$:
\begin{equation*}
\begin{split}
  &\left(f_1\left(Y_{1j}\right),\dots,f_p\left(Y_{pj}\right)\right)\sim\mbox{\bf{mN}}_{p}\left(\boldsymbol{0}_{p},\bTheta_{p\times p}^{-1}\right),
\end{split}
\end{equation*}
where $\bTheta_{p\times p}$ is an unknown precision matrix. In this case $Y_j=\left(Y_{1j},\dots,Y_{pj}\right)$ is said to follow a nonparanormal multivariate normal distribution, $Y_j\sim\mbox{\bf{NPN}}\left(\boldsymbol{0}_{p},\bTheta_{p\times p}^{-1},f\right)$. Then they inferred the precision matrix $\bTheta_{p\times p}$ with the following objective function from \textit{graphical lasso} \cite{friedman2008sparse}:
\[\min_{\bTheta_{p\times p}}\left\{\mbox{tr}\left(\bTheta_{p\times p}\mathbf{S}\right)-\log|\bTheta_{p\times p}|+\beta\sum_{i_1,i_2}\bTheta_{i_1i_2}\right\},\]
where $\mathbf{S}$ is the empirical covariance matrix of $\left(f_1\left(Y_{1j}\right),\dots,f_p\left(Y_{pj}\right)\right),\;j=1,\dots,n$ in \textit{graphical lasso}, and $\beta$ is the regularization parameter controlling sparsity. \cite{liu2012high} used the estimated correlation matrix $\hat{\mathbf{S}}$ instead of $\mathbf{S}$, estimated using Kendall's tau or Spearman's rho. In particular, one defines $\Delta_{i}(j,j') = Y_{i j} - Y_{ij'} $, so that
\vspace{-5pt}
\begin{equation*}
    \begin{split}
       &\text{(Kendall's tau)    }\\ &\hat{\tau}_{i_1i_2}=\frac{2}{n\left(n-1\right)}\sum_{j<j^{'}}\mbox{sign}\left(\Delta_{i_1}(j,j')\Delta_{i_2}(j,j')\right),    
    \end{split}
\end{equation*}
\vspace{-5pt}
\begin{equation*}
    \begin{split}
     &\text{(Spearman's rho)}\\ &\hat{\rho}_{i_1i_2}=\frac{\sum_{j=1}^n\left(r_{i_1j}^{\left(c\right)}-\bar{r}_j^{\left(c\right)}\right)\left(r_{i_2j}^{\left(c\right)}-\bar{r}_j^{\left(c\right)}\right)}{\sqrt{\sum_{j=1}^n\left(r_{i_1j}^{\left(c\right)}-\bar{r}_j^{\left(c\right)}\right)^2\left(r_{i_2j}^{\left(c\right)}-\bar{r}_j^{\left(c\right)}\right)^2}},  
    \end{split}
\end{equation*}
where $r_{ij}^{(c)}$ is the rank of $Y_{ij}$ among $Y_{1j},\dots,Y_{pj}$ and $\bar{r}^{(c)}_j=\frac{1}{p}\sum_{i=1}^pr_{ij}^{(c)}=\frac{1+p}{2}$.
Correspondingly,
\begin{equation*}
   \hat{\mathbf{S}}_{i_1i_2}=
    \begin{cases}
      \sin\left(\frac{\pi}{2}\hat{\tau}^{\left(c\right)}_{i_1i_2}\right), & i_1\neq i_2,\\
      1, & i_1=i_2.
    \end{cases}       
\end{equation*}
\begin{equation*}
   \hat{\mathbf{S}}_{i_1i_2}=
   \begin{cases}
      2\sin\left(\frac{\pi}{6}\hat{\rho}^{\left(c\right)}_{i_1i_2}\right), & i_1\neq i_2,\\
     1, & i_1=i_2.
    \end{cases} 
\end{equation*}
\cite{ning2013high} extended the matrix-normal distribution with Kronecker product structure to non-Gaussian data with a similar semiparametric approach applied on both the row vectors and the column vectors of $\bf{Y}$.

\subsection{Background on Bigraphical lasso}
Bigraphical Lasso is introduced by \cite{kalaitzis2013bigraphical}. Let $\mathbf{Y} \in  \mathbb{R}^{n \times p}$ be a random matrix. If its rows are generated as i.i.d. samples from $N \left(0, \bSigma_{p\times p}\right)$, then the sampling distribution of the sufficient statistic $\mathbf{Y}^\top\mathbf{Y} $ is $\text{Wishart}\left(n, \bSigma_{p\times p}\right)$. At the same time, if the columns are generated as $i.i.d.$ samples from $\mathcal{N}\left(0, \bGamma_{p\times p}\right)$, then the sampling distribution is $\text{Wishart}\left(n, \bGamma_{p\times p}\right)$.
Combining these sufficient statistics in a model for the entire matrix $\mathbf{Y}$ as
\begin{equation*}
p\left(\mathbf{Y}\right)\propto\exp\{-\mbox{tr}\left(\bPsi_{n\times n} \mathbf{Y}\mathbf{Y}^\top\right) -\mbox{tr}\left(\bTheta_{p\times p}\mathbf{Y}^\top\mathbf{Y}\right)\}
\end{equation*}
%
is equivalent to a joint factorised Gaussian distribution for the entries of $\mathbf{Y}$, with a precision matrix given by the $KS$:
\begin{equation}
\bOmega=\bPsi_{n\times n}\oplus\bTheta_{p\times p} = \bPsi_{n\times n} \otimes \mathbf{I}_p + \mathbf{I}_n \otimes \bTheta_{p\times p}.
\label{eq:ks}
\end{equation}
Through this representation we obtain a parameter vector of size
$O\left(n^2 +p^2\right)$ instead of the usual $O\left(n^2p^2\right)$.

Given data in the form of some design matrix $\mathbf{Y}$, the Bigraphical Lasso model proposed in \cite{kalaitzis2013bigraphical} estimates the sparse $KS$-structured inverse covariance of a matrix normal by minimising the $\ell_1$-penalized negative likelihood function of ($\bPsi_{n\times n}$ , $\bTheta_{p\times p}$):
\begin{equation}
  \label{eq:objective}
  \begin{split}
  \min_{\bTheta_{p\times p},\bPsi_{n\times n}}\bigg\{n \mbox{tr}\left(\bTheta_{p\times p}\mathbf{S}\right)+p \mbox{tr}\left(\bPsi_{n\times n}\mathbf{T}\right)-\log| \bPsi_{n\times n} \oplus \bTheta_{p\times p}|
 &+ \beta_1 ||\bPsi_{n\times n}||_1 +\beta_2 ||\bTheta_{p\times p}||_1 \bigg\},
  \end{split}
\end{equation}
\par\noindent where $\mathbf{S} \triangleq \tfrac{1}{n}\mathbf{Y}^\top\mathbf{Y}$ and $\mathbf{T} \triangleq \tfrac{1}{p}\mathbf{Y}\mathbf{Y}^\top$ are empirical covariances across the samples and features respectively.
    A solution simultaneously estimates two graphs --- one over the columns of $\mathbf{Y}$, corresponding to the sparsity pattern of $\bTheta_{p\times p}$, and another over the rows of $\mathbf{Y}$, corresponding to the sparsity pattern of $\bPsi_{n\times n}$.

The original paper of \cite{kalaitzis2013bigraphical}  proposes a \textit{flip-flop} approach first optimizing over $\bPsi_{n\times n}$, while holding $\bTheta_{p\times p}$ fixed, and then optimizing over $\bTheta_{p\times p}$ while holding $\bPsi_{n\times n}$ fixed. They show that in case of no regularization, the first step of the optimization problem is reduced to
\begin{equation*} 
  \underset{\bPsi_{n\times n}}{\textrm{min}} \Big\{ p\mbox{tr}\left(\bPsi_{n\times n} \mathbf{T}\right) - \ln \verts{\bPsi_{n\times n} \oplus \bTheta_{p\times p}} \Big\}.
\end{equation*}
Obtaining the stationary point:
\begin{equation} \label{eq:stationary}
  \mathbf{T} - \tfrac{1}{2p} \mathbf{T} \circ \mathbf{I} = \tfrac{1}{p}\textrm{tr}_p \left(\mathbf{W}\right) - \tfrac{1}{2p}\textrm{tr}_p \left(\mathbf{W}\right) \circ \mathbf{I} ~,
\end{equation}
where $\circ$ is the Hadamard product and we define $\mathbf{W} \triangleq \left(\bPsi_{n\times n} \oplus \bTheta_{p\times p}\right)^{-1}$. The block-wise trace $\text{tr}_p\left(\cdot\right)$ is an operator that  to each $np \times np$ matrix $\mathbf{M}$ written in terms of $n^2$ many $p\times p$ blocks
\[
\mathbf{M} = \begin{bmatrix}
  \mathbf{M}_{11} &\dots &\mathbf{M}_{1n} \\
  \vdots &\ddots &\vdots \\
  \mathbf{M}_{n1} &\dots &\mathbf{M}_{nn}
\end{bmatrix},
\]
associates the matrix of traces of each $p\times p$ block:
  \[
  \textrm{tr}_p\left(\mathbf{M}\right) =
  \begin{bmatrix}
    \mbox{tr}\left(\mathbf{M}_{11}\right)&\dots &\mbox{tr}\left(\mathbf{M}_{1n}\right)\\ \vdots &\ddots &\vdots \\ \mbox{tr}\left(\mathbf{M}_{n1}\right) &\dots &\mbox{tr}\left(\mathbf{M}_{nn}\right)
    \end{bmatrix},
    \]
as defined in  \cite{kalaitzis2013bigraphical}.
While their approach dramatically reduces the computational complexity
of the problem, its memory requirements (i.e. space complexity) are prohibitive for problems
involving large $n$ or $p$. 

Our contribution in Section 3 is to give a more efficient solution in terms of computational and space complexity.
\section{SCALABLE BIGRAPHICAL LASSO ALGORITHM}
Consider the eigen-decomposition of the two precision matrices $\bPsi_{n\times n} =\mathbf{U}\Lambda_1\mathbf{U}^\top$ and $\bTheta_{p\times p}=\mathbf{V}\Lambda_2\mathbf{V}^\top$, where $\Lambda_1\in \mathbb{R}^{n\times n}$ and $\Lambda_2\in \mathbb{R}^{p\times p}$ are  eigenvalues diagonal matrices and $\mathbf{U}=\left(u_{ij}\right)\in \mathbb{R}^n$ and $\mathbf{V}=\left(v_{ij}\right)\in \mathbb{R}^p$ are orthogonal eigenvectors matrices associated respectively to $\bPsi_{n\times n}$ and $\bTheta_{p\times p}$. It follows that Equation (\ref{eq:ks}) can be rewritten as
\begin{align}\label{Omega_eigenDecompo}
  \bOmega=\left(\mathbf{U}\otimes \mathbf{V}\right) [\Lambda_1\otimes \mathbf{I}_p + \mathbf{I}_n \otimes \Lambda_2] \left(\mathbf{U}^\top \otimes \mathbf{V}^\top\right).
\end{align}
Inversion of a symmetric matrix for which an eigenvalue decomposition is provided is achieved through inversion of the eigenvalues,
\begin{equation*}
\mathbf{W} =\bOmega^{-1}= \left(\mathbf{U}\otimes \mathbf{V}\right) [\Lambda_1\otimes \mathbf{I}_p + \mathbf{I}_n \otimes \Lambda_2] ^{-1}\left(\mathbf{U}^\top \otimes \mathbf{V}^\top\right).
\end{equation*}
Taking
\begin{equation*}
\left(\mathbf{I}_n\otimes \mathbf{V}^\top\right) \left(\mathbf{I}_n\otimes \mathbf{I}_p\right)=  \mathbf{I}_n \otimes \mathbf{V}^\top,
\end{equation*}
then 
\begin{equation}
\mathbf{W} \bOmega= \mathbf{I}_n\otimes \mathbf{I}_p
\label{eq:womega}
\end{equation}
can be premultiplied by $\mathbf{I}_n \otimes \mathbf{V}^\top$ to provide
\begin{eqnarray}
\left(\mathbf{I}_n\otimes \mathbf{V}^\top\right) \mathbf{W} \bOmega=\left(\mathbf{U}\otimes \mathbf{I}_p\right)\mathbf{D}\left(\mathbf{U}^\top \otimes \mathbf{V}^\top\right)\bOmega,
\label{eq:wup}
\end{eqnarray}
where $\mathbf{D}= [\Lambda_1\otimes \mathbf{I}_p + \mathbf{I}_n
\otimes \Lambda_2] ^{-1}$ is a diagonal matrix. 
The detailed proof of Eq. (\ref{Omega_eigenDecompo}) and Eq. (\ref{eq:wup}) can be found in the Supplementary Material A.1.
%
%
%
%
%
Multiply both sides of Equation (\ref{eq:wup}) by $\mathbf{I}_n \otimes \mathbf{V}$, we have
\begin{equation} \label{eq:womegaexp}
    \mathbf{I}_n \otimes \mathbf{I}_p=\left(\mathbf{U}\otimes \mathbf{I}_p\right)\mathbf{D}\left(\mathbf{U}^\top \otimes \mathbf{I}_p\right)\left(\bPsi_{n\times n}\otimes \mathbf{I}_p+ \mathbf{I}_n \otimes \Lambda_2\right),
\end{equation}
Detailed proof of Eq. (\ref{eq:womegaexp}) can be found in the Supplementary Material A.2.
%
Eq (\ref{eq:womegaexp}) can be rewritten in a similar form as Equation (\ref{eq:womega})
\begin{equation*} 
  \hat{\mathbf{W}}\hat{\mathbf{\bOmega}} = \mathbf{I}_n \otimes \mathbf{I}_p,
\end{equation*}
where
\begin{equation*}
  \hat{\mathbf{W}}=  \left[\mathbf{U}\otimes \mathbf{I}_p\right]\mathbf{D}\left[\mathbf{U}^\top \otimes \mathbf{I}_p\right]
\end{equation*}
and
\begin{equation*}
  \hat{\mathbf{\bOmega}}= \bPsi_{n\times n}\otimes \mathbf{I}_p + \mathbf{I}_n \otimes \Lambda_2.
\end{equation*}
We partition $\hat{\mathbf{W}}$ and $\hat{\mathbf{\bOmega}}$ into blocks
\[
\hat{\mathbf{W}} =
\begin{bmatrix}
  \hat{\mathbf{W}} _{11} &\hat{\mathbf{W}}_{1 \sm1} \\ \hat{\mathbf{W}}_{\sm1 1} & \hat{\mathbf{W}} _{\sm1 \sm1}
\end{bmatrix},
\]
\[
\hat{\bOmega} = \begin{bmatrix}
  \hat{\bOmega}_{11} &\hat{\mathbf{\bOmega}}_{1\sm1} \\ \hat{\mathbf{\bOmega}}_{\sm1 1} & \hat{\mathbf{\bOmega}}_{\sm1 \sm1}
\end{bmatrix},
\]
where $\hat{\mathbf{W}}_{11}$ and $\hat{\mathbf{\bOmega}}_{11}$ are $p \times p$ matrices and $\hat{\mathbf{W}}_{\sm1 1}$ and $\hat{\mathbf{\bOmega}}_{\sm1 1}$ are $p\left(n-1\right) \times p$ matrices. Then from the bottom-left block of
\begin{equation} \label{equ:WOmega}
    \hat{\mathbf{W}}\hat{\mathbf{\bOmega}}
    = \hat{\mathbf{W}}\left(\bPsi_{n\times n}\otimes\mathbf{I}_p+\mathbf{I}_n\otimes\Lambda_2\right)=\mathbf{I}_{n} \otimes \mathbf{I}_p
\end{equation}
we get
\begin{equation*}
  \hat{\mathbf{W}}_{\sm1 1} \left(\psi_{11}\mathbf{I}_p + \Lambda_2\right) + \hat{\mathbf{W}}_{\sm1\sm1} \left(\bpsi_{\sm1 1} \otimes \mathbf{I}_p\right)=\mathbf{0}_{n-1} \otimes \mathbf{I}_p,
\end{equation*}
where we use notation $\bPsi_{n\times n}=\left(\psi_{ij}\right)_{i,j=1,\dots,n}$ and $\bpsi_{\sm1 1}$ represents the corresponding sub-block.
Post multiplying both sides of the last equation by $\left(\psi_{11}\mathbf{I}_p + \Lambda_2\right)^{-1}$ we have
\begin{equation} \label{eq:W1_not1}
  \hat{\mathbf{W}}_{\sm1 1} + \hat{\mathbf{W}}_{\sm1\sm1}
  \begin{bmatrix}
    \left(\psi_{11}\mathbf{I}_p + \Lambda_2\right)^{-1} \psi_{21} \\
    \vdots \\
    \left(\psi_{11}\mathbf{I}_p + \Lambda_2\right)^{-1}\psi_{n1}
  \end{bmatrix}
  = \mathbf{0}_{n-1} \otimes \mathbf{I}_p.
\end{equation}
Detailed proof of Eq. (\ref{eq:W1_not1}) can be found in the Supplementary Material A.3.
\par Decomposing $\hat{\mathbf{W}}_{\sm1\sm1}$ in $\left(n-1\right)$ adjacent blocks $\hat{\mathbf{W}}_{\sm1 k} \in \mathbb{R}^{\left(n-1\right)p\times p}$, $\forall k\in \left\{2,\ldots, n\right\}$, then Equation (\ref{eq:W1_not1}) can be rewritten as
\begin{equation*}
  \begin{split}
    \hat{\mathbf{W}}_{\sm1 1} + \hat{\mathbf{W}}_{\sm1 2} \left(\psi_{11}\mathbf{I}_p + \Lambda_2\right)^{-1} \psi_{21} +\dots+ \hat{\mathbf{W}}_{\sm1 n} \left(\psi_{11}\mathbf{I}_p + \Lambda_2\right)^{-1} \psi_{n1} = \mathbf{0}_{n-1} \otimes \mathbf{I}_p.
  \end{split}
\end{equation*}

\paragraph{Proposition 3.1} Following the assumptions and calculations above we have \begin{equation*}
    \text{tr}_p\left(\mathbf{W}\right)= \text{tr}_p\left(\hat{\mathbf{W}}\right).
\end{equation*}
The proof of Proposition 3.1 is in the Supplementary Material.
Proposition 3.1 enables us to make use of the stationary point given in Equation \eqref{eq:stationary}. 
As described in \cite{kalaitzis2013bigraphical}, we can partition the empirical covariance $ \mathbf{T}$ as
\[
\mathbf{T} =
\begin{bmatrix}
  \mathbf{t} _{11} &\mathbf{t}_{1 \sm1} \\ \mathbf{t}_{\sm1 1} & \mathbf{T} _{\sm1 \sm1}
\end{bmatrix},
\]
where $\mathbf{t}_{ \sm1 1} \in \mathbb{R}^{n-1}$ and  $\mathbf{T} _{\sm1 \sm1} \in \mathbb{R}^{\left(n-1\right)\times\left(n-1\right)}$. In particular, from the lower left block of \eqref{eq:stationary} we get
\begin{equation*}
  \mathbf{t}_{ \sm1 1}=\frac{1}{p}\text{tr}_p\left(\mathbf{W}_{\sm1 1}\right).
\end{equation*}  

Taking the block-wise trace $\textrm{tr}_p\left(\cdot\right)$ of both sides of \eqref{eq:W1_not1}, gives
\begin{equation}
  p  \mathbf{t}_{\sm1 1} + \mathbf{A}_{\sm 1 \sm1} \bpsi_{\sm1 1} = \mathbf{0}_{n-1},
  \label{eq:lasso}
\end{equation}
where $\mathbf{A}^\top_{\sm 1 \sm1} \in \mathbb{R}^{\left(n-1\right) \times \left(n-1\right)}$ is:
\begin{equation}
  \mathbf{A}^\top_{\sm 1 \sm1} \triangleq
  \begin{bmatrix}\textrm{tr}_p \left\{ \hat{\mathbf{W}}_{\sm1 2} \left(\psi_{11}\mathbf{I}_p + \bLambda_2\right)^{-1} \right\}^\top \\
    \vdots \\
    \textrm{tr}_p \left\{ \hat{\mathbf{W}}_{\sm1 n} \left(\psi_{11}\mathbf{I}_p + \bLambda_2\right)^{-1} \right\}^\top
  \end{bmatrix}.
  \label{eq:A}
\end{equation}

The problem posed in Equation \eqref{eq:lasso} is addressed via a lasso regression. In Proposition 3.2 
we use some of the previous decomposition in order to reduce the computational complexity of the problem.
\vspace{-5pt}
\paragraph{Proposition 3.2} Following the assumptions and calculations above we have
\begin{align}\label{eqn:trace}
\begin{split}
&\textrm{tr}_p\left\{ \hat{\mathbf{W}}_{\sm1 k} \left(\psi_{11}\mathbf{I}_p + \bLambda_2\right)^{-1} \right\}=
     \sum_{j=1}^p \frac{1}{\psi_{11}+\lambda_{2j}}
     \nonumber  \begin{bmatrix}
       \sum_{i=1}^n \frac{u_{2i}u_{ki}}{\lambda_{1i}+\lambda_{21}}\\
       \vdots& \\
       \sum_{i=1}^n \frac{u_{ni}u_{ki}}{\lambda_{1i}+\lambda_{2p}}
     \end{bmatrix},
     \end{split}
\end{align}
where  $\lambda_{11} \ldots \lambda_{1n}$ and $\lambda_{21} \ldots \lambda_{2p}$ are the diagonal values of $\Lambda_1\in \mathbb{R}^{n\times n}$ and $\Lambda_2 \in \mathbb{R}^{p \times p}$, respectively.
The proof of Proposition 3.2 is in the Supplementary Material.
\par We note that by imposing an $\ell_1$ penalty on $\bPsi_{\sm1 1}$, the problem posed in (\ref{eq:lasso}) reduces to a lasso regression involving now only the matrix ${\bf U}$, the diagonal of ${\boldsymbol \Lambda}_1$ and ${\boldsymbol \Lambda}_2$, and $\psi_{11}$. This decomposition frees the prohibitive amount of memory needed to store the matrix $\hat{\mathbf{W}}$, which is of size $n^2p^2$. 

\par The lasso regression will provide an estimation on the first column of $\bPsi_{n\times n}$. For the update of all the other columns $\bPsi_{\sm i i}$ we need to reiterate the same approach. Indeed we  partition $\bPsi_{n\times n}$ into $\psi_{ii}, \bpsi_{\sm i i}$ and $\bPsi_{\sm i \sm i}$ for $i=1,\ldots,n$. We then find a sparse solution of $ p  \mathbf{t}_{\sm i i} + \mathbf{A}_{\sm i \sm i} \bpsi_{\sm i i} = \mathbf{0}_{n-1}$ with \emph{lasso} regression. Given the new value $\bpsi_{\sm i i} $ we then compute the eigenvalues matrix ${\boldsymbol \Lambda}_1$ and eigenvectors matrix ${\bf U}$ of $\bPsi_{n\times n}$. This will provide the updated values to be used in Proposition 3.2. Hence, after $n$ steps, the columns of $\bPsi_{n\times n}$ are estimated. Similarly the estimation of $\bTheta_{p\times p}$, for fixed $\bPsi_{n\times n}$, becomes directly analogous to the above simply by \emph{transposing} the design matrix (samples become features and vice-versa) and is obtained in $p$ steps.  In our experiments 
the precision matrices $\bPsi_{n\times n}$ and $\bTheta_{p\times p}$ are initialised as identity matrices. The empirical mean matrix is removed from each dataset.

\setlength{\textfloatsep}{5pt}
\begin{algorithm}[h]
  \caption{scBiGLasso} 
  \begin{algorithmic}
    \State\textbf{Input:} Maximum iteration number $N$, tolerance $\varepsilon$, $m$ many observations of $p\times n$ matrices $\mb{Y}^{(k)}$, $k=1,\dots,m$. $\beta_1, \beta_2$ and initial estimates of $\bPsi_{n\times n}$ and $\bTheta_{p\times p}$, $\bPsi_{n\times n}^{(0)}$ and $\bTheta_{p\times p}^{(0)}$.
    \State For each $\mb{Y}^{(k)}$, $\mb{T^{(k)}} \leftarrow p^{-1}\mb{Y^{(k)}}\mb{Y^{(k)}}^\top$.
    \State $\hat{\mathbf{T}}\leftarrow \frac{1}{m}\sum_{k=1}^{m}\hat{\mathbf{T}}^{(k)}$
    \Repeat
    \State \# \emph{Estimate $\bPsi_{n\times n}$}~:
    \For {iteration $\tau=1,\dots,N$}
    \For{$i = 1,\dots,n$}
    \State Partition $\bPsi^{(\tau-1)}_{n\times n}$ into $\psi^{(\tau-1)}_{ii}, \bpsi^{(\tau-1)}_{i \sm i}$ and \\
    $~$\hspace{1.5cm}$\bPsi^{(\tau-1)}_{\sm i \sm i}$.
    \State Calculate $\mb{A}^{(\tau-1)}_{\sm i \sm i} $ as in Equation (\ref{eq:A}) with\\ $~$\hspace{1.5cm}$\psi^{(\tau-1)}_{ii}$.
    \State With \emph{Lasso} regression, find a sparse \\$~$\hspace{1.5cm}solution of
    $p\mb{t}_{i \sm i} +\mb{A}^{(\tau-1)}_{\sm i \sm i}\bpsi^{(\tau)}_{i \sm i} =\mb{0}_{n-1}$. 
    \State Update the eigen-decomposition of the\\
    $~$\hspace{1.4cm} precision matrix $\bPsi^{(\tau)}_{n\times n} =\mathbf{U}{\boldsymbol \Lambda}_1\mathbf{U}^\top$ 
    \EndFor
    \State \# \emph{Estimate $\bTheta_{p\times p}$}~:
    \State Proceed as if estimating $\bPsi_{n\times n}$ with input\\
    $~$\hspace{1cm}$\mb{Y}^\top, \beta_1, \beta_2$.
    \State \[\Delta\bPsi^{(\tau)}=\|\bPsi^{(\tau)}_{n\times n}-\bPsi^{(\tau-1)}_{n\times n}\|_F^2\]
    \[\Delta\bTheta^{(\tau)}=\|\bTheta^{(\tau)}_{p\times p}-\bTheta^{(\tau-1)}_{p\times p}\|_F^2\]
    \EndFor
    \Until {Maximum iteration number reached, or\\ $~$\hspace{-0.25cm}$\underset{\tau^*=\tau-2,\tau-1,\tau}{\max}\left\{(\Delta\bPsi^{(\tau^*)}+\Delta\bTheta^{(\tau^*)})\right\}<\varepsilon$, for $\tau\geq 3$. }
  \end{algorithmic}
\end{algorithm}

\par The approach is summarised in Algorithm 1 for Gaussian data. We point out that the convergence of Algorithm 1 could also be
directly verified on the value of the objective function
(\ref{eq:objective}) at each step, but, due to the computation of $|
\bPsi_{n\times n} \oplus \bTheta_{p\times p}|$ , when $p,n>>100$ this becomes
unfeasible. Indeed, the space complexity can be reduced from
$O\left(n^2p^2\right)$ to $O\left(n^2+p^2\right)$ by means of Proposition 3.3.

\paragraph{Proposition 3.3} Following the assumptions and calculations above we have
\begin{equation*}
    | \bPsi_{n\times n} \oplus \bTheta_{p\times p} | =\nonumber \prod_{i=1}^n\prod_{j=1}^p\left(\lambda_{1i}+\lambda_{2j}\right).
\end{equation*}
The proof of Proposition 3.3 is in the Supplementary Material.

It follows that:
\vspace{-2em}
\begin{equation*}
\log | \bPsi_{n\times n} \oplus \bTheta_{p\times p} | = \sum_{i=1}^n\sum_{j=1}^p\log|\lambda_{1i}+\lambda_{2j}|=K.
\end{equation*}

Hence we can write the objective function as
\begin{equation*}
\begin{split}
\min_{\bTheta_{p\times p},\bPsi_{n\times n}}\bigg\{n \mbox{tr}\left(\bTheta_{p\times p}\mathbf{S}\right)+p \mbox{tr}\left(\bPsi_{n\times n}\mathbf{T}\right) - K
& + \beta_1 ||\bPsi_{n\times n}||_1 +\beta_2 ||\bTheta_{p\times p}||_1 \bigg\}.  
\end{split}
\end{equation*}
\vspace{-10pt}


Note that this scalable version of the Bigraphical Lasso enables
higher dimensional problems. This is mainly due to the fact that in
our implementation there is no need to directly evaluate the matrix $\bf{W}$. Instead we just need the eigen-decomposition of the  two precision matrices $\bPsi_{n\times n}$ and $\bTheta_{p\times p}$. In the original paper \cite{kalaitzis2013bigraphical} at each step $i$ the blocks of $\mathbf{W}$ are explicitly updated and of course were involved in the next step of the estimation. In particular $\mathbf{W}_{\sm i i}$ is computed via backward-substitution in Equation (\ref{eq:W1_not1}) and $W_{11}$ via backward-substitution in Equation (\ref{equ:WOmega}). 
\par In summary, as we are not interested in the estimation of the overall $\hat{\mathbf{W}}$ nor $\boldsymbol{\Omega}$, we will never explicitly update them, but we will rather focus on the estimation of $\bPsi_{n\times n}$ and $\bTheta_{p\times p}$. This leads to a space complexity reduction from $O(n^2p^2)$ to $O(n^2 + p^2)$ by means of Proposition 3.2 and Proposition 3.3.
\begin{figure}[H]
\vspace{-0.5em}
    \begin{center}
        \subfloat[\label{genworkflow}]{\includegraphics[trim=100 250 120 270,clip, width=0.422\textwidth]
{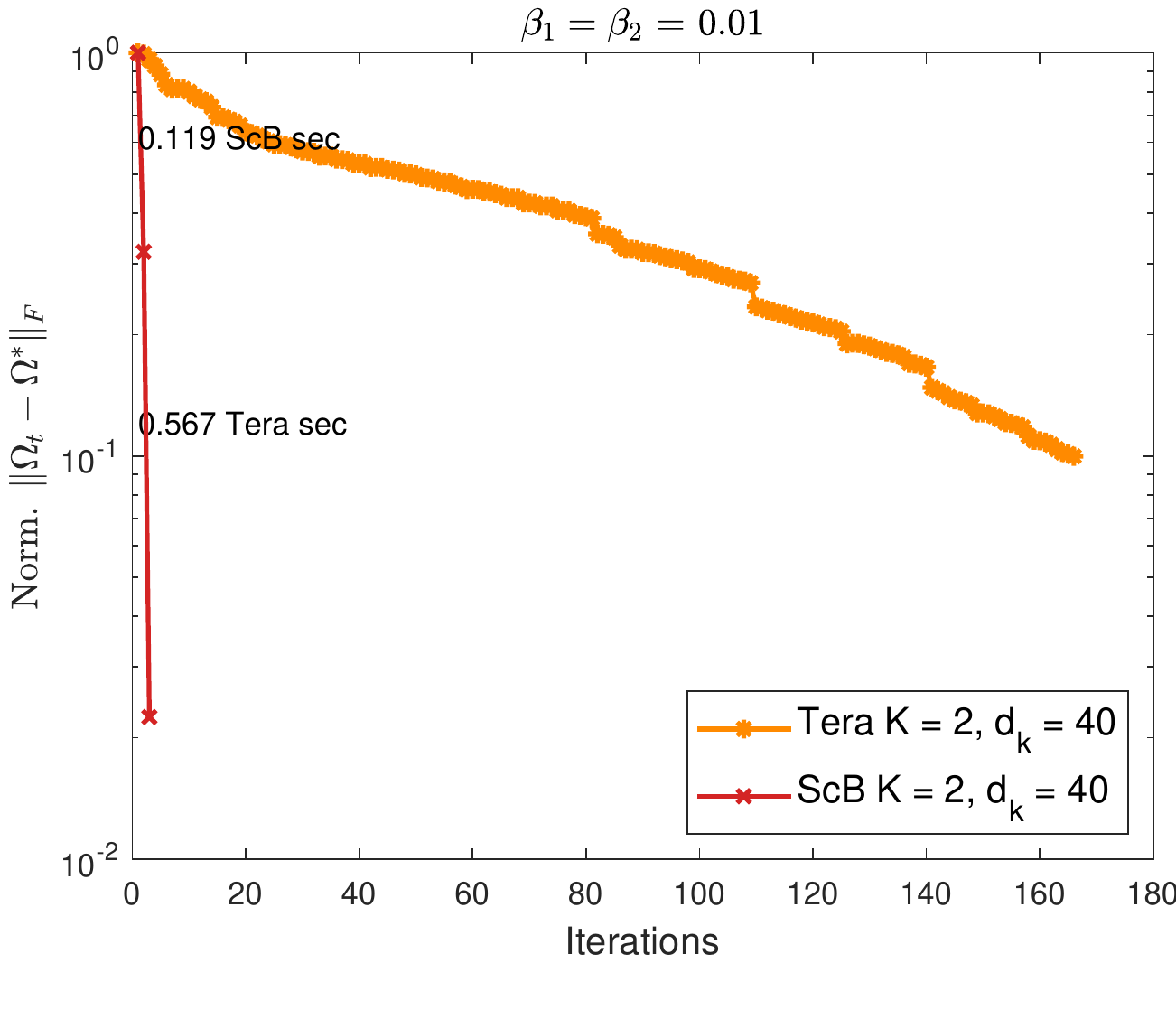}}
  \subfloat[\label{pyramidprocess} ]{\includegraphics[trim=120 250 120 270,clip, width=0.4\textwidth]
{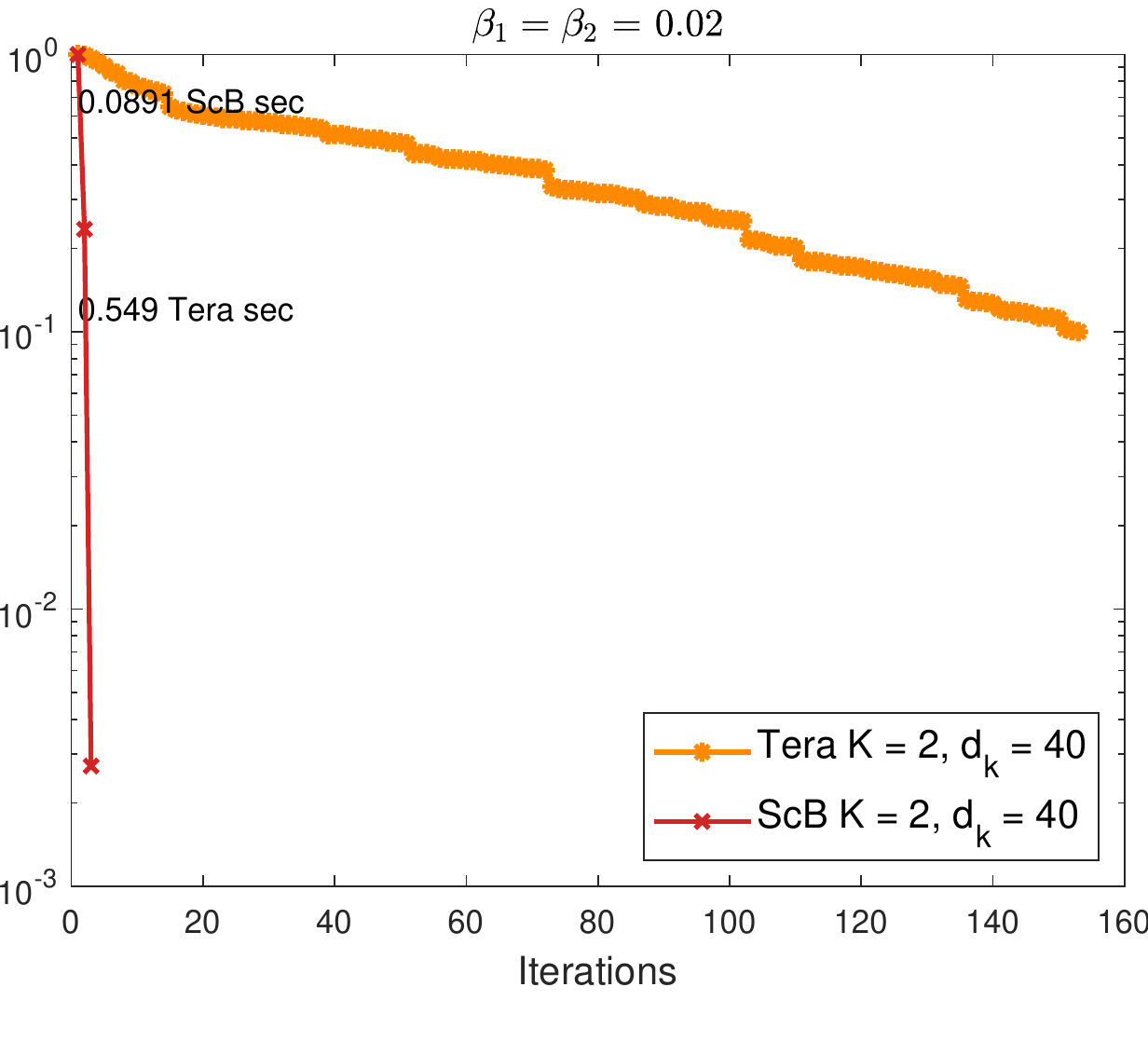}}
    \caption{\label{workflow}ScB and Tera convergence rates and times with regularisation parameters $\beta_1=\beta_2\in\{0.01,0.02\}$.}
    \label{rebuttalFigure}
  \end{center}
\end{figure}
\vspace{-2em}
\par Our model provides a Scalable Bigraphical lasso algorithm (ScB) and as such benefits of the same statistical convergence properties. A subgaussian concentration inequality \cite[Lemma 19, Supplementary Material]{greenewald2019tensor} gives rates of statistical convergence \cite[Theorems 1-3]{greenewald2019tensor} of the TeraLasso estimator as well as the Bigraphical Lasso estimator, when the sample size is low. 
In Figure \ref{rebuttalFigure} we show the numerical convergence rates and times of ScB with respect to the Frobenius norm for the precision matrix, compared to the Teralasso approach with K=2. 

\section{NONPARANORMAL BIGRAPHICAL MODEL}
The method in Section 3 only deals with Gaussian data, while in real world many data come in the form of count data. In this section, we introduce a Gaussian copula based method to adapt Algorithm 1 for count data.
We start with the definition of the matrix nonparanormal distribution with a Kronecker sum structure.
\paragraph{Definition 4.1}
Consider a  $p\times n$ non-Gaussian data matrix $\mathbf{Y}$. $\mathbf{Y}$ follows a matrix nonparanormal 
distribution with a Kronecker sum structure $\mbox{\bf{MNPN}}_{KS}\left(\mathbf{M};\mathbf{\bPsi}_{n\times n}^{-1},\mathbf{\bTheta}_{p\times p}^{-1};f\right)$, with mean matrix $\mathbf{M}$, and where $\bPsi_{n\times n}$ and $\bTheta_{p\times p}$ are the row-specific and the column-specific precision matrices, if and only if there exists a set of monotonic transformations $f=\left\{f_{ij}\right\}_{i=1,\dots,p}^{j=1,\dots,n}$ such that
\begin{equation*}
   \mbox{vec}{\left[f\left(\mathbf{Y}\right)\right]}
   \sim \mbox{\bf mN}\left(\mbox{vec}{\left(\mathbf{M}\right)},\left(\bPsi_{n\times n}\oplus\bTheta_{p\times p}\right)^{-1}\right). 
   \vspace{-5pt}
\end{equation*}
In this paper, we only consider the model after centering, i.e $\mbox{vec}({\mathbf{M}})=\boldsymbol{0}_{np}$. The choices $f_{ij}\left(Y_{ij}\right)=Y_{ij}$ and $f_{ij}\left(Y_{ij}\right)=\log{Y_{ij}}$ give us multivariate Normal distribution and multivariate log-Normal distribution respectively. Since we only require $f$ to be monotone, this model provides us with a wider family of distributions to work on, thus extends the Bigraphical model to non-Gaussian data. We note that the model in Definition 4.1 can be viewed as a latent model, with latent variable $\mathbf{Z}=f\left(\mathbf{Y}\right)$ and $\mbox{vec}\left(\mathbf{Z}\right)\sim\mbox{\bf{mN}}\left(\boldsymbol{0}_{np},\left(\bPsi_{n\times n}\oplus\bTheta_{p\times p}\right)^{-1}\right)$.
\par Following the arguments in \cite{kalaitzis2013bigraphical} and \cite{greenewald2019tensor}, the supports of $\bPsi_{n\times n}$ and $\bTheta_{p\times p}$ encode the dependence structure of the row variables and the column variables, respectively. Further discussion and mathematical details of the decomposition of the latent model are in the Supplementary Material A.7.
\par In the next section, we introduce a method to infer the nonparanormal distribution without explicitly defining $f$.
\subsection{Estimation of the precision matrices}
We now consider estimation of the precision matrices $\bPsi_{n\times n}$ and $\bTheta_{p\times p}$. Like the lasso methods applied in one-way network inference and in Gaussian Bigraphical models, we enforce sparsity on $\bPsi_{n\times n}$ and $\bTheta_{p\times p}$ by regularization on the negative log-likelihood, which gives us the objective function:
\begin{equation*}
\begin{split}
  \min_{\bPsi_{n\times n},\bTheta_{p\times p}} &\bigg\{p\mbox{tr}\left(\bPsi_{n\times n}\mathbf{T}\right)+n\mbox{tr}\left(\bTheta_{p\times p}\mathbf{S}\right)-K
  +\beta_1\|\bPsi_{n\times n}\|_{1}
   +\beta_2\|\bTheta_{p\times p}\|_1\bigg\},
\end{split}
\end{equation*}
where $\mathbf{T}=\frac{1}{p}\left(\bf{Z}\bf{Z}^{\top}\right)$ is the empirical covariance matrix along the rows, and $\mathbf{S}=\frac{1}{n}\left(\bf{Z}^\top\bf{Z}\right)$ is the empirical covariance matrix along the columns.
The only problem that remains now is to estimate the empirical covariance matrices $\mathbf{T}$ and $\mathbf{S}$. When estimating one-way network, \cite{liu2012high} proposed the nonparanormal skeptic, exploiting Kendall's tau or Spearman's rho, without explicitly calculating the marginal transforming function $f$. Similarly, we define Kendall's tau and Spearman's rho along rows and columns. More specifically, let $r_{ij}^{\left(c\right)}$ be the rank of $Y_{ij}$ among $Y_{1j},\dots,Y_{pj}$ and $\bar{r}_j^{\left(c\right)}=\frac{1}{p}\sum_{i=1}^{p}r_{ij}=\frac{p+1}{2}$. Define $\Delta_{i}(j,j') = Y_{i j} - Y_{ij'} $. We consider the following statistics:
\begin{equation*}
    \begin{split}
       &\text{(Column-wise Kendall's tau)    }\\ &\hat{\tau}^{\left(c\right)}_{i_1i_2}=\frac{2}{n\left(n-1\right)}\sum_{j<j^{'}}\mbox{sign}\left(\Delta_{i_1}(j,j')\Delta_{i_2}(j,j')\right),    
    \end{split}
\end{equation*}
\begin{equation*}
    \begin{split}
      &\text{(Column-wise Spearman's rho)}\\ &\hat{\rho}^{\left(c\right)}_{i_1i_2}=\frac{\sum_{j=1}^n\left(r_{i_1j}^{\left(c\right)}-\bar{r}_j^{\left(c\right)}\right)\left(r_{i_2j}^{\left(c\right)}-\bar{r}_j^{\left(c\right)}\right)}{\sqrt{\sum_{j=1}^n\left(r_{i_1j}^{\left(c\right)}-\bar{r}_j^{\left(c\right)}\right)^2\left(r_{i_2j}^{\left(c\right)}-\bar{r}_j^{\left(c\right)}\right)^2}}.  
    \end{split}
\end{equation*}
Similarly, let $r_{ij}^{\left(r\right)}$ be the rank of $Y_{ij}$ among $Y_{i1},\dots,Y_{in}$ and $\bar{r}_i^{\left(r\right)}=\frac{1}{n}\sum_{j=1}^{n}r_{ij}=\frac{n+1}{2}$. Define $\Delta_{j}(i,i') = Y_{i j} - Y_{i'j} $. We consider the following statistics:
\begin{equation*}
\begin{split}
  &\text{(Row-wise Kendall's tau)}\\ &\hat{\tau}^{\left(r\right)}_{j_1j_2}=\frac{2}{p\left(p-1\right)}\sum_{i<i^{'}}\mbox{sign}\left(\Delta_{j_1}(i,i')\Delta_{j_2}(i,i')\right),
  \end{split}
\end{equation*}
\begin{equation*}
\begin{split}
   &\text{(Row-wise Spearman's rho)}\\ &\hat{\rho}^{\left(r\right)}_{j_1j_2}=\frac{\sum_{i=1}^p\left(r_{ij_1}^{\left(r\right)}-\bar{r}_i^{\left(r\right)}\right)\left(r_{ij_2}^{\left(r\right)}-\bar{r}_i^{\left(r\right)}\right)}{\sqrt{\sum_{i=1}^p\left(r_{ij_1}^{\left(r\right)}-\bar{r}_i^{\left(r\right)}\right)^2\left(r_{ij_2}^{\left(r\right)}-\bar{r}_i^{\left(r\right)}\right)^2}}.
\end{split}
\end{equation*}
And the following estimated covariance matrices using Kendall's tau and Spearman's rho:
\begin{equation}\label{T_hat_Kendall}
   \hat{\mathbf{T}}_{j_1j_2}=
    \begin{cases}
      \sin\left(\frac{\pi}{2}\hat{\tau}^{\left(r\right)}_{j_1j_2}\right), & j_1\neq j_2,\\
      1, & j_1=j_2.
    \end{cases}       
\end{equation}
\begin{equation}\label{T_hat_Spearman}
   \hat{\mathbf{T}}_{j_1j_2}=
    \begin{cases}
      2\sin\left(\frac{\pi}{6}\hat{\rho}^{\left(r\right)}_{j_1j_2}\right), & j_1\neq j_2,\\
      1, & j_1=j_2.
    \end{cases} 
\end{equation}
\begin{equation*}
   \hat{\mathbf{S}}_{i_1i_2}=
    \begin{cases}
      \sin\left(\frac{\pi}{2}\hat{\tau}^{\left(c\right)}_{i_1i_2}\right), & i_1\neq i_2,\\
      1, & i_1=i_2.
    \end{cases}       
\end{equation*}
\begin{equation*}
   \hat{\mathbf{S}}_{i_1i_2}=
    \begin{cases}
      2\sin\left(\frac{\pi}{6}\hat{\rho}^{\left(c\right)}_{i_1i_2}\right), & i_1\neq i_2,\\
      1, & i_1=i_2.
    \end{cases} 
\end{equation*}
In Algorithm 2 we summarise the Nonparanormal Scalable Bigraphical Lasso approach for count data. 
\setlength{\textfloatsep}{10pt}
\begin{algorithm}[H]\label{nonparanormal_scb}
  \caption{Nonparanormal scBiGLasso} 
  \begin{algorithmic}
    \State\textbf{Input:} Maximum iteration number $N$, tolerance $\varepsilon$, $m$ many observations of $p\times n$ count matrices $\mb{Y}^{(k)}$, $k=1,\dots,m$. $\beta_1, \beta_2$ and initial estimates of $\bPsi_{n\times n}$ and $\bTheta_{p\times p}$, $\bPsi_{n\times n}^{(0)}$ and $\bTheta_{p\times p}^{(0)}$.
    \State For each $\mb{Y}^{(k)}$, calculate $\hat{\mathbf{T}}^{(k)}$ according to Equation (\ref{T_hat_Kendall}) or (\ref{T_hat_Spearman}).
    \State $\hat{\mathbf{T}}\leftarrow \frac{1}{m}\sum_{k=1}^{m}\hat{\mathbf{T}}^{(k)}$
    \Repeat
    \State \# \emph{Estimate $\bPsi_{n\times n}$}~:
    \For {iteration $\tau=1,\dots,N$}
    \For{$i = 1,\dots,n$}
    \State Partition $\bPsi^{(\tau-1)}_{n\times n}$ into $\psi^{(\tau-1)}_{ii}, \bpsi^{(\tau-1)}_{i \sm i}$ and \\
    $~$\hspace{1.5cm}$\bPsi^{(\tau-1)}_{\sm i \sm i}$.
    \State Calculate $\mb{A}^{(\tau-1)}_{\sm i \sm i} $ as in Equation (\ref{eq:A}) with\\ $~$\hspace{1.5cm}$\psi^{(\tau-1)}_{ii}$.
    \State With \emph{Lasso} regression, find a sparse \\$~$\hspace{1.5cm}solution of
    $p\mb{t}_{i \sm i} +\mb{A}^{(\tau-1)}_{\sm i \sm i}\bpsi^{(\tau)}_{i \sm i} =\mb{0}_{n-1}$. 
    \State Update the eigen-decomposition of the\\
    $~$\hspace{1.4cm} precision matrix $\bPsi^{(\tau)}_{n\times n} =\mathbf{U}{\boldsymbol \Lambda}_1\mathbf{U}^\top$ 
    \EndFor
    \State \# \emph{Estimate $\bTheta_{p\times p}$}~:
    \State Proceed as if estimating $\bPsi_{n\times n}$ with input\\
    $~$\hspace{1cm}$\mb{Y}^\top, \beta_1, \beta_2$.
    \State \[\Delta\bPsi^{(\tau)}=\|\bPsi^{(\tau)}_{n\times n}-\bPsi^{(\tau-1)}_{n\times n}\|_F^2\]
    \[\Delta\bTheta^{(\tau)}=\|\bTheta^{(\tau)}_{p\times p}-\bTheta^{(\tau-1)}_{p\times p}\|_F^2\]
    \EndFor
    \Until {Maximum iteration number reached, or\\ $~$\hspace{-0.25cm}$\underset{\tau^*=\tau-2,\tau-1,\tau}{\max}\left\{(\Delta\bPsi^{(\tau^*)}+\Delta\bTheta^{(\tau^*)})\right\}<\varepsilon$, for $\tau\geq 3$. }
  \end{algorithmic}
\end{algorithm}
\section{NUMERICAL RESULTS}
In this Section, we implement our Scalable Bigraphical Lasso algorithm where covariance matrices are estimated using Kendall's tau. After precision matrices $\bPsi_{n\times n}$ and $\bTheta_{p\times p}$ are inferred, they are transformed into binary matrices to reveal the network structures, where any non-zero value in the precision matrices become $1$ and any zero value stays zero. We illustrate an application of our overall approach on both synthetic and real datasets as described in the following subsections.
\subsection{Synthetic Gaussian Data}
To demonstrate the efficiency of our Scalable Bigraphical Lasso algorithm (Algorithm 1), we generate sparse positive definite matrices $\bPsi_{n\times n}$ and $\bTheta_{p\times p}$. Then simulate $m$ many $p\times n$ Gaussian data $Y_{G}^{(k)},\;k=1,\dots,m$ from $\mbox{mN}\left(\boldsymbol{0},\left(\bPsi_{n\times n}\oplus\bTheta_{p\times p}\right)^{-1}\right)$. We plug $Y_{G}^{(k)},\;k=1,\dots,m$ into our implemented Algorithm 1, Bigraphical Lasso from \cite{kalaitzis2013bigraphical} and TeraLasso from \cite{greenewald2019tensor}. Figure 1 shows a comparison between the convergence times of Algorithm 1 and Bigraphical Lasso for increasing problem dimensions $n=p$. We can observe that, as expected, Algorithm 1 converges in significantly faster times, allowing one to tackle higher dimensional problems in practice. Table 1 shows the network recovery when $n=p=100$. We can see that our method provides high Accuracy while improving greatly on speed; see Section 5.2 for the definition of Accuracy.

\begin{figure}[h]
    \begin{center}
        \includegraphics[width=0.6\textwidth]{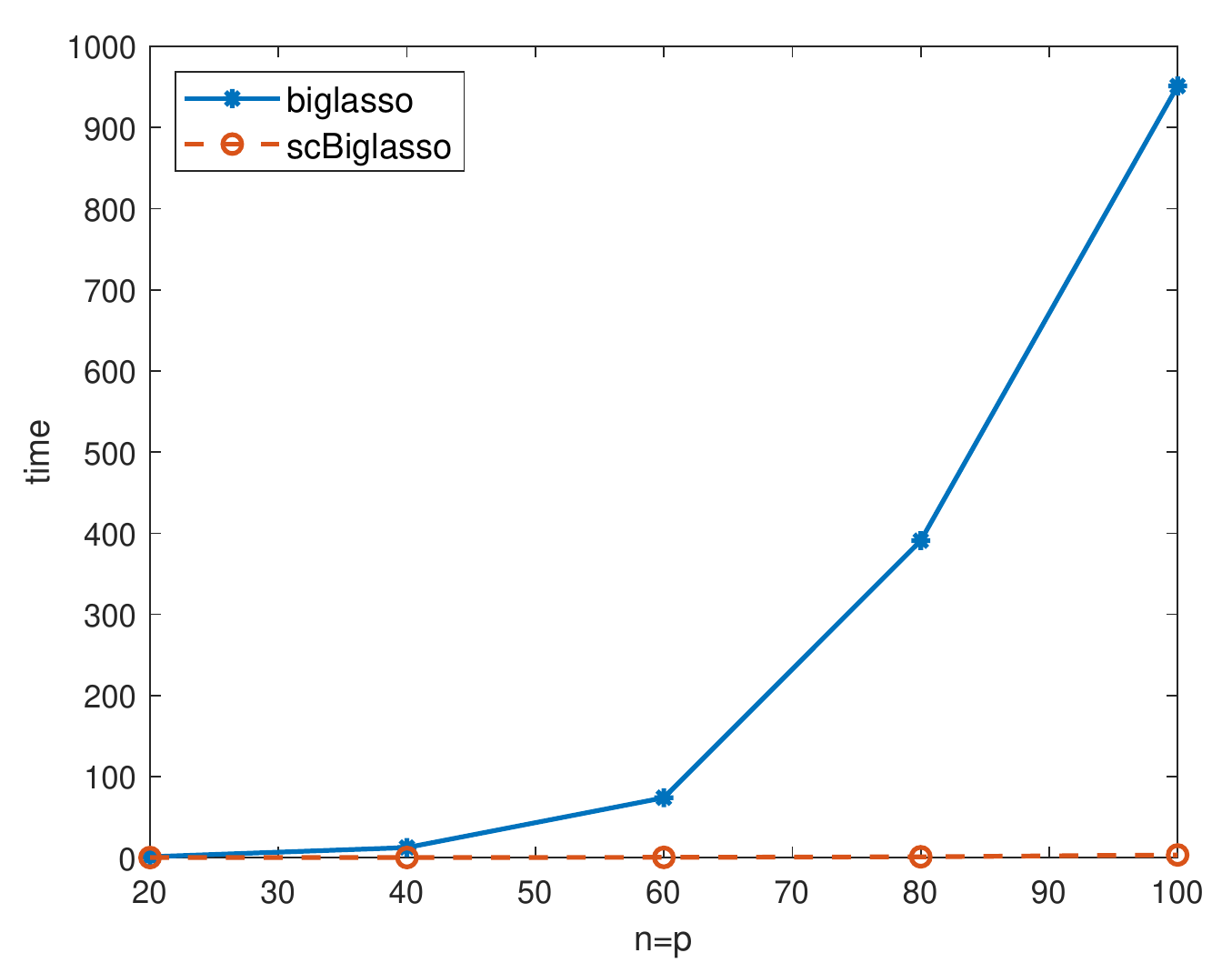}%
    \caption{Computational convergence time ($seconds$) comparison between Bigraphical Lasso (\cite{kalaitzis2013bigraphical}) and Algorithm 1, for increasing values of the dataset dimensions $n=p$.}
    \label{synth_gaussian}
  \end{center}
\end{figure}
\vspace{-2.3em}
\begin{table}[H]
\caption{Comparison between computational convergence times, Accuracy of $\bPsi$ and of $\bTheta$ for Bigraphical Lasso (\cite{kalaitzis2013bigraphical}), TeraLasso (\cite{greenewald2019tensor}) and Algorithm 1, for a synthetic Gaussian dataset with dimensions $n=p=100$.} \label{Tab:1}
\centering
\begin{tabular}{llll}
\textbf{Method}  &\textbf{Accuracy$_\Psi$}
&\textbf{Accuracy$_\Theta$}&\textbf{Time(s)}\\
\hline \\
Biglasso         &$0.9032$ & $0.9028$ & $951.15$\\
ScBiglasso            & $0.9032$ & $0.9028$ & $3.50$\\
TeraLasso            & $0.5416$ & $0.4323$ & $0.3696$ \\
\end{tabular}
\end{table}


\vspace{-5pt}
\subsection{Synthetic count data}
\vspace{-5pt}
We generate and process Gaussian Copula-based count data through the following steps:
\vspace{-10pt}
\begin{enumerate}
\item Generate sparse positive definite matrix $\bPsi_{n\times n}$ and $\bTheta_{p\times p}$. Calculate the Kronecker sum of $\bPsi_{n\times n}$ and $\bTheta_{p\times p}$.Generate $m$ multivariate-normal vectors of length $p\times n$ from $\mbox{mN}\left(\boldsymbol{0},\bOmega^{-1}\right)$, where $\bOmega=\bPsi_{n\times n}\otimes I_p+I_n\otimes\bTheta_{p\times p}$. 
\item Centre each of the $m$ multivariate-normal vectors around their mean, and reshape the vectors into $p\times n$ matrices $X^{\left(1\right)},\dots,X^{\left(m\right)}$.
\item For each $X^{\left(k\right)}$, $k=1,\dots,m$, calculate the matrix $P^{\left(k\right)}$ such that $P_{ij}^{(k)}=\Phi\left(X_{ij}^{(k)}\right)$, where $\Phi\left(\cdot\right)$ is the cumulative density function of the standard normal distribution.
\item For each $k=1,\dots,m$, produce the negative binomial variable $Y_{ij}^{\left(k\right)}=QNB\left(P_{ij}^{\left(k\right)},r,p\right)$, where $QNB\left(\cdot,r,p\right)$ is the quantile function of $\mbox{Negative-Binomial}\left(r,p\right)$, with $r$ the number of success to be observed and $p$ the success rate.
\end{enumerate}

\par Below we describe some of the criteria we use to assess the recovery of the synthetic network. Denote $TP$ as the number of \textit{True Positives} in the network recovery, $TN$ as the number of \textit{True Negatives} in the network recovery, $FP$ as the number of \textit{False Positives} in the network recovery, and $FN$ the number of \textit{False Negatives} in the network recovery, then we can define
\begin{eqnarray*}
&&Precision=\frac{TP}{TP+FP},\quad Recall=\frac{TP}{TP+FN},\\
&&Accuracy=\frac{TP+TN}{TP+TN+FP+FN},\\
&&TPR=\frac{TP}{TP+FN},\quad FPR=\frac{FP}{TN+FP}.
\end{eqnarray*}
\par
\begin{figure}[h]
    \begin{center}
        \includegraphics[width=0.9\textwidth]
        {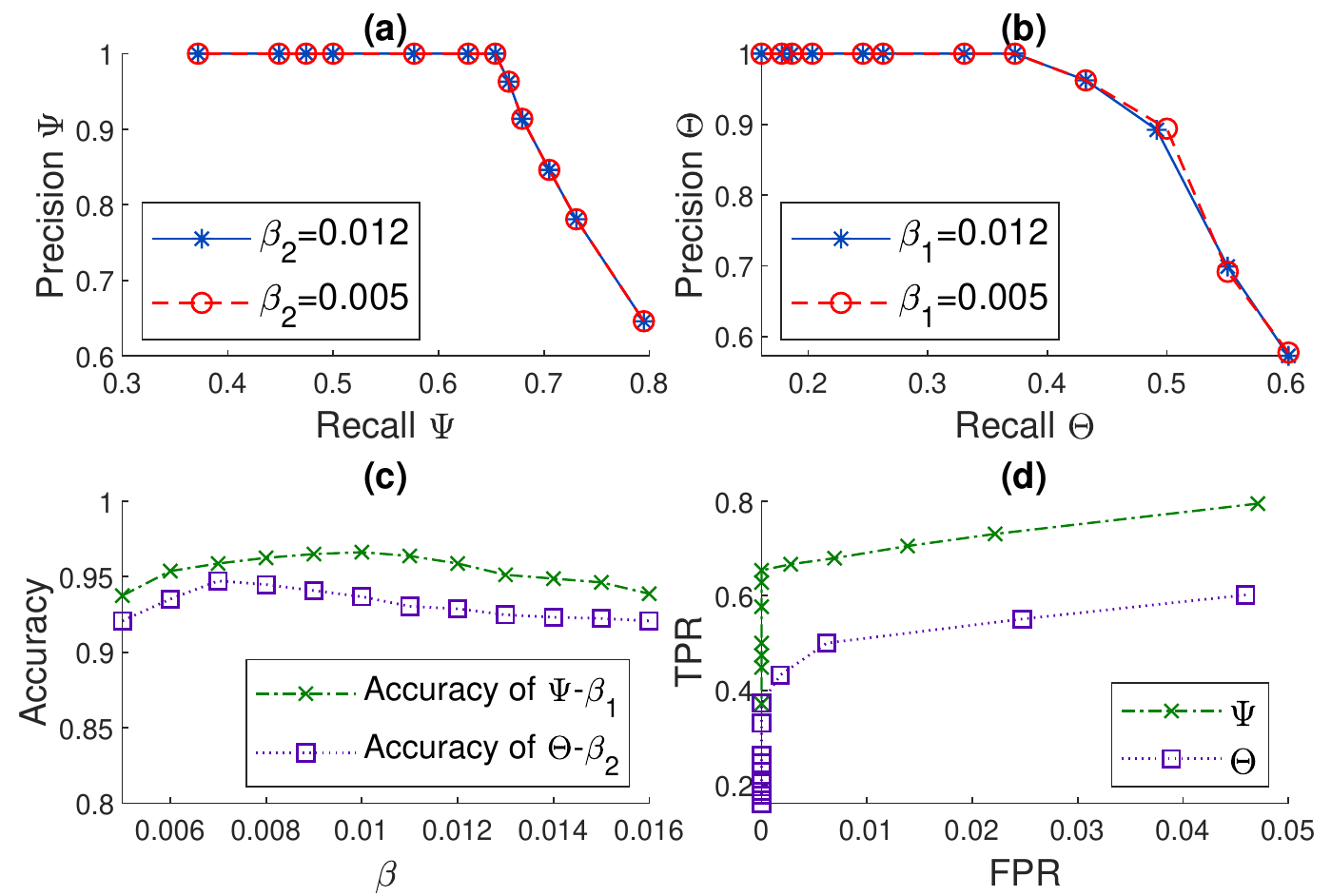}
        \vspace{-12pt}
    \caption{Synthetic network recovery results. {\bf (a)} Precision-Recall of the network recovery relating to the support of $\bPsi_{n\times n}$; {\bf (b)} Precision-Recall of the network recovery relating to the support of $\bTheta_{p\times p}$; {\bf (c)} Accuracy vs corresponding regularization parameter $\beta_1$ ($\beta_2$) of the network recovery relating to the support of $\bPsi_{n\times n}$ ($\bTheta_{p\times p}$) and {\bf (d)} TPR-FPR of the network recovery relating to the support of $\bPsi_{n\times n}$ ($\bTheta_{p\times p}$), where the corresponding regularization parameter $\beta_1$ ($\beta_2$) $\in\left\{0.005:0.001:0.0016\right\}$.}
    \label{fig:synth_result}
    \end{center}
\end{figure}

 Figure \ref{fig:synth_result} shows some results from synthetic data. Figure \ref{fig:synth_result} (a) is the Precision-Recall of the recovery of $\bPsi_{n\times n}$ with changing $\beta_1$ (different points on the graph) and $\beta_2$ (different colours on the graph). Two arbitrary values of $\beta_2$ have been chosen to illustrate how the results do not depend on $\beta_2$. This is expected as $\beta_1$ is the regularization parameter for $\bPsi_{n\times n}$, while $\beta_2$ corresponds to $\bTheta_{p\times p}$. A similar result is shown in Figure \ref{fig:synth_result} (b), where the Precision-Recall of the recovery of $\bTheta_{p\times p}$ heavily depends on the choice of $\beta_2$, regardless of the $\beta_1$ value. Figure \ref{fig:synth_result} (c)(d) show that high values of $TPR$ and Accuracy, with low values of $FPR$, can be achieved for appropriate choices of $\beta_1$ and $\beta_2$ in the range $\left[0.005,0.016\right]$.
\par
\begin{figure}[H]
    \begin{center}
        \includegraphics[width=0.8\textwidth]
        {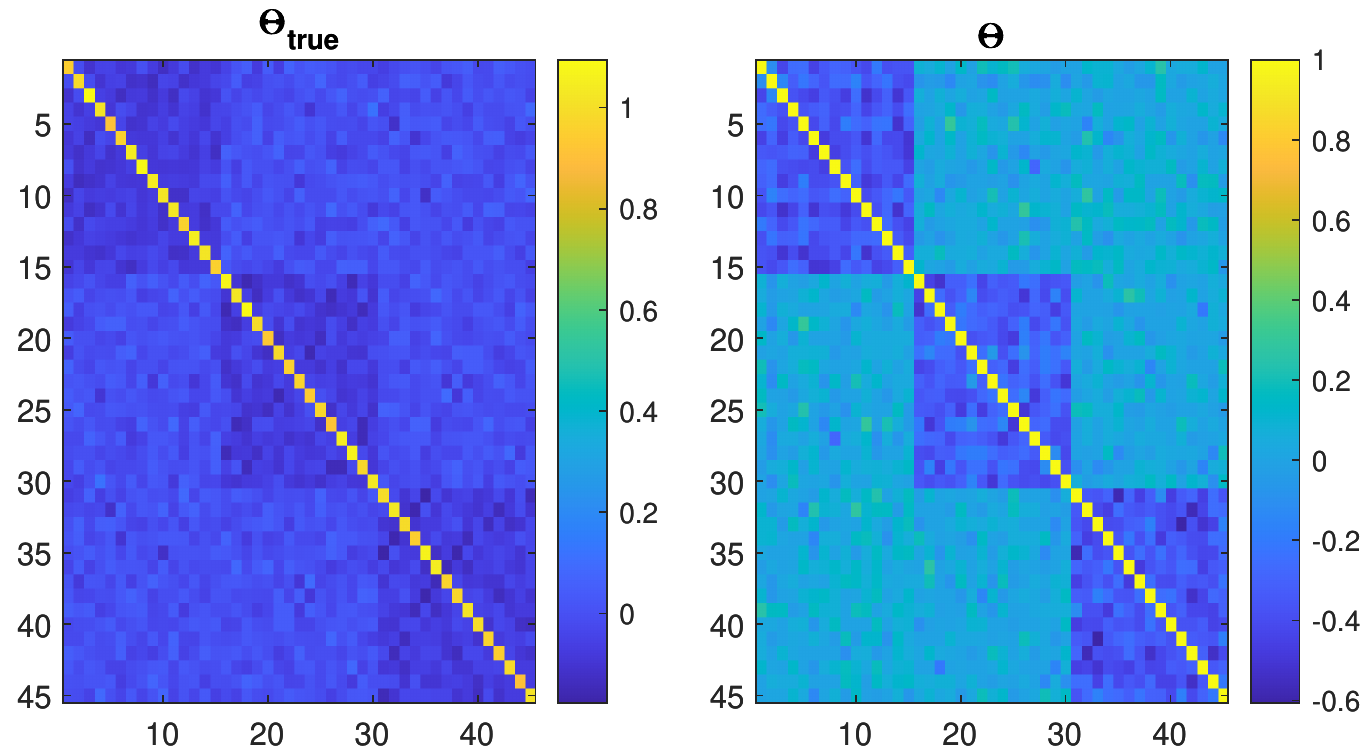}
        \vspace{-5pt}
    \caption{Synthetic network recovery. We generated synthetic data as described in Section $5.2$ using a block-diagonal precision matrix for $\bTheta_0$ plus Gaussian noise (Left plot). On the right we plot the estimated $\bTheta$ via our method. In this example, we used  $\beta_2=0.0002$.}
    \label{synth_block}
  \end{center}
\end{figure}

 Figure \ref{synth_block} shows network recovery for another synthetic count dataset, where the original precision matrix $\bTheta_0$ was generated with block diagonals and Gaussian noise throughout the matrix. We observe that our method leads to good recovery of the corresponding blocks. Further discussion on the choice of optimal regularization parameters $\beta=(\beta_1,\beta_2)$ is in the Supplementary Material.
\vspace{-5pt}


\vspace{-10pt}

\subsection{mESC scRNA-seq data}
We use a single cell gene expression dataset from mouse embryonic stem cells (mESC)  available in \cite{buettner2015computational}. The data consist of measurements of gene counts in 182 single cells at different stages of the cell cycle. We will refer to the three phases as G1, S and G2M. About $700$ genes are annotated as cell cycle related. Of these, we considered $167$ genes more active during mitosis, the cell division phase and last part of the cell cycle (G2M). In our dataset there are 65 cells in the G2M phase. 
\par
\begin{figure}[h]
    \begin{center}
      \includegraphics[width=0.8\textwidth]{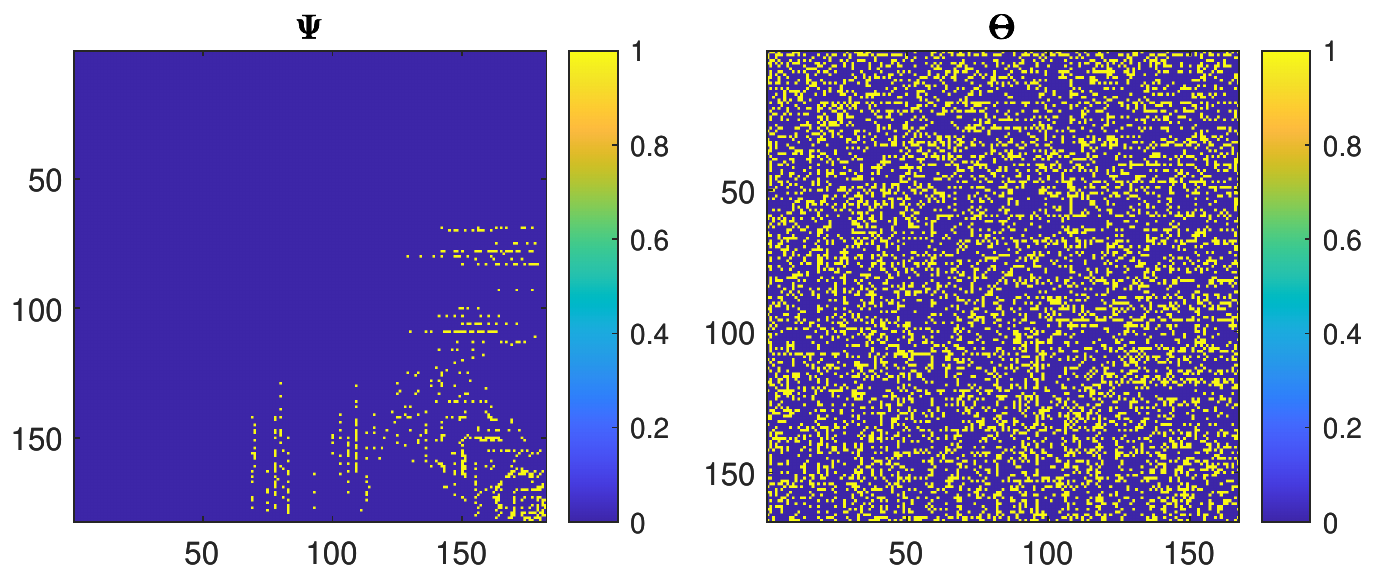}
    \caption{Networks recovered by our proposed Scalable Bigraphical Lasso algorithm combined with the nonparanormal transformation as described in Section 4.2, $\left(\beta_1,\beta_2\right)=\left(0.014,0.001\right)$.}
    \label{fig:real_theta_result}
    \end{center}
\end{figure}

\begin{figure}[H]
\vspace{-5pt}
\centering
\includegraphics[width=0.4\textwidth]{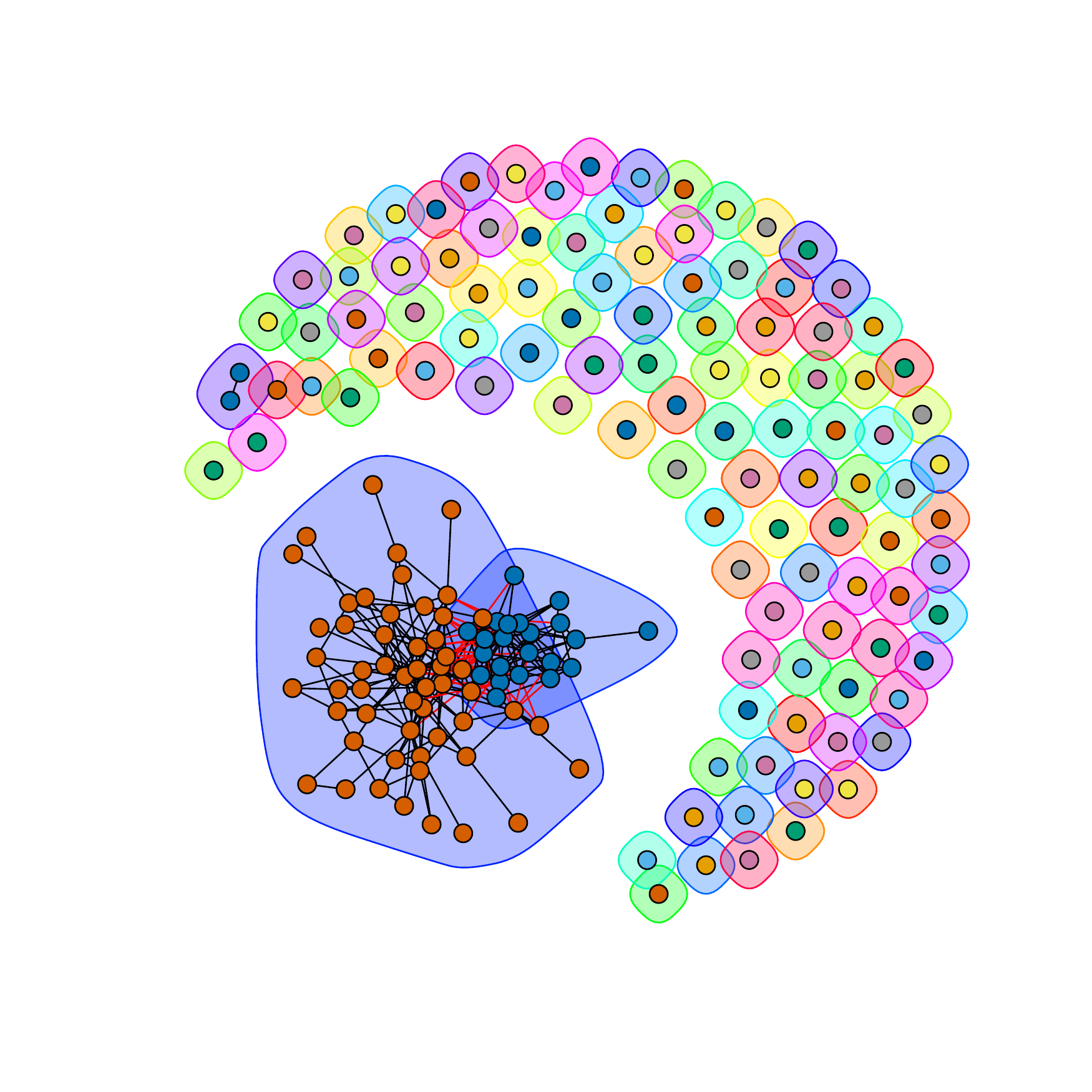}
\includegraphics[width=0.4\textwidth]{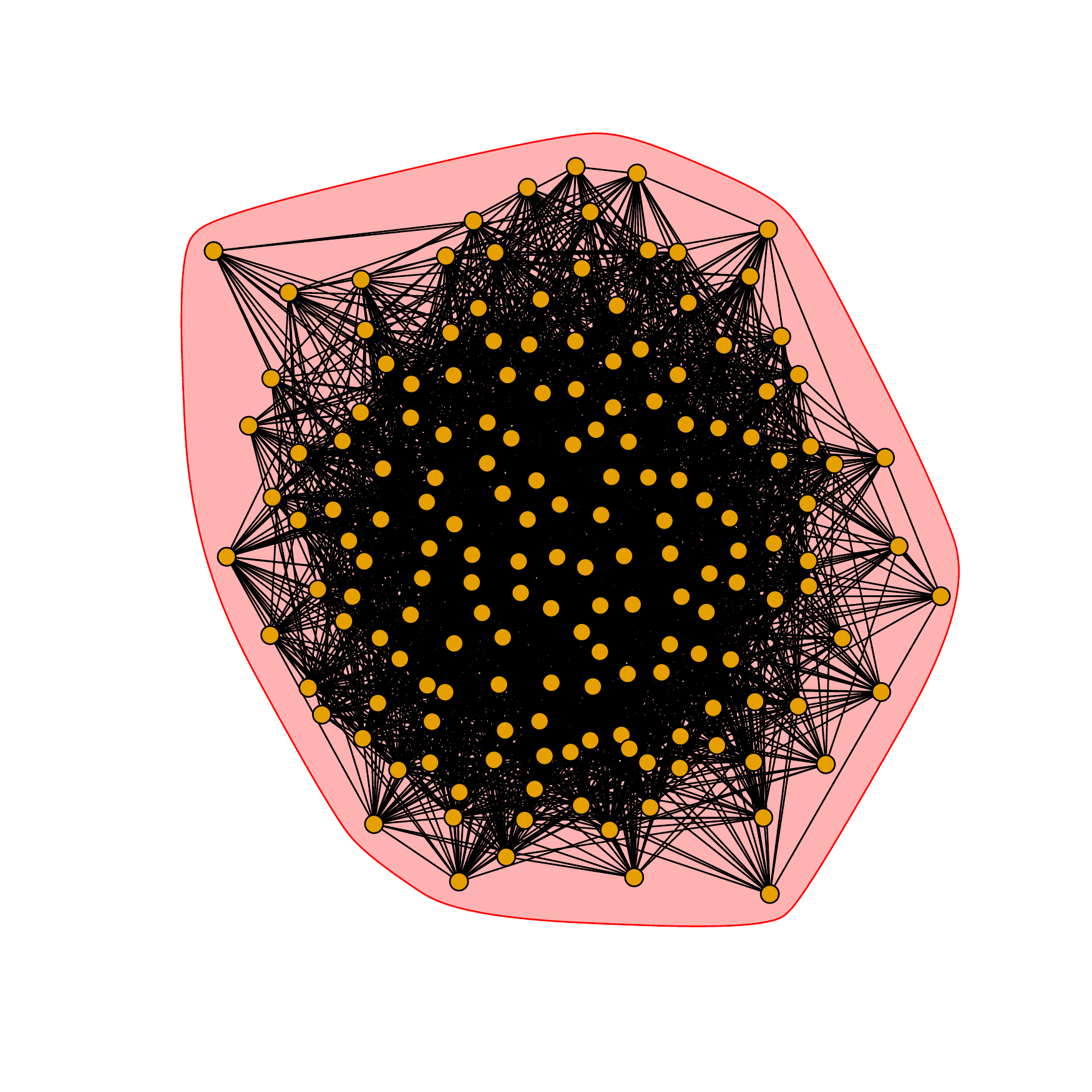}
\vspace{-10pt}
\caption{$\bPsi$ (left) and $\bTheta$ (right) induced networks.}
\label{realgraph}
\end{figure}

 In Figures \ref{fig:real_theta_result} and \ref{realgraph}, we show how our model allows the identification of the sub-population of cells that correspond to the G2M stage. In Figure \ref{fig:real_theta_result} we show the estimated precision matrices for the cells (left) and the genes (right). We use a binary transformation where only the negative values are considered an edge in the network. In Figure \ref{realgraph} we plot the corresponding networks, over imposing the clusters found with the label propagation approach developed by \cite{raghavan2007near}. We note that $\sim 92\%$ of the G2M cells are clustered in two densely connected modules ($\bPsi$ network plot in Figure \ref{realgraph}), while no connection is measured between cells in different phases of the cell cycle. As expected, on the other hand, the mitosis genes are all densely connected in a single cluster ($\bTheta$ network plot in Figure \ref{realgraph}).    

\vspace{-10pt}

\section{CONCLUSIONS}
In this work, we present a Scalable Bigraphical Lasso algorithm. In particular, we exploit eigenvalue decomposition of the Cartesian product graph to present a more efficient version of the algorithm presented in \cite{kalaitzis2013bigraphical}. Our approach reduces memory requirements from $O(n^2p^2)$ to $O(n^2 + p^2)$, and reduces the computational time by up to a factor of 200 in our experiments (case $p=n=100$ in Figure \ref{synth_gaussian} and Table \ref{Tab:1}). Note that comparisons for $n=p>100$ were restricted because of the memory limitation in \cite{kalaitzis2013bigraphical}. Additionally, we propose a Gaussian-copula based model and a semiparametric approach that enables the application of the proposed Bigraphical model to non-Gaussian data. This is particularly relevant for count data applications, such as single cell data. Future work will include optimisation of the choice of the regularization parameters, and potential extension to $k$-way network inference for non-Gaussian data, with $k>2$.
\subsubsection*{Data availability}

The code and data is available at \url{https://github.com/luisacutillo78/Scalable_Bigraphical_Lasso.git}.

\subsubsection*{Acknowledgements}
Sijia Li was supported by an EPSRC Doctoral Training Partnership (reference EP/R513258/1) through the University of Leeds. 
The authors would like to thank Michael Croucher for the support in optimizing the MATLAB code. Sijia Li would like to thank Nicole M\"{u}cke for the mentorship on the writing. Luisa Cutillo and Neil Lawrence would like to acknowledge the Marie Curie fellowship CONTESSA (ID: 660388), during which the main ideas of this research were conceived. The authors would like to thank the reviewers and editor for their constructive criticism of the manuscript. 
\printbibliography
\newpage
\section*{Supplementary Material}
\subsection*{A Mathematical analysis}
\par In this Section, we provide detailed proofs for some of the properties and results in the main paper.

\subsubsection*{A.1 Proof of Equations (\ref{Omega_eigenDecompo}) and (\ref{eq:wup})}
Equation (\ref{Omega_eigenDecompo}) in the main paper follows from the following:
\begin{equation*}
  \bOmega \ = \ \bPsi_{n\times n}\oplus\bTheta_{p\times p} 
  \ = \ \mathbf{U}\Lambda_1\mathbf{U}^\top\otimes \mathbf{I}_p + \mathbf{I}_n \otimes \mathbf{V}\Lambda_2\mathbf{V}^\top
  \ = \ \left(\mathbf{U}\otimes \mathbf{V}\right) [\Lambda_1\otimes \mathbf{I}_p + \mathbf{I}_n \otimes \Lambda_2] \left(\mathbf{U}^\top \otimes \mathbf{V}^\top\right).
\end{equation*}

%
Equation (\ref{eq:wup}) within the main paper follows from Equation (\ref{Omega_eigenDecompo}). In particular, we have
\begin{equation*}\label{eq1split}
\begin{split}
  \left(\mathbf{U}\otimes \mathbf{I}_p\right)\mathbf{D}\left(\mathbf{U}^\top \otimes \mathbf{V}^\top\right)\bOmega
  &=  \left(\mathbf{U}\otimes \mathbf{I}_p\right)\mathbf{D}\left(\mathbf{U}^\top \otimes \mathbf{V}^\top\right)\left( \bPsi_{n\times n} \otimes \mathbf{I}_p + \mathbf{I}_n \otimes \mathbf{V}\Lambda_2\mathbf{V}^\top\right)\\
  &=\left(\mathbf{U}\otimes \mathbf{I}_p\right)\mathbf{D}\left(\mathbf{U}^\top \otimes \mathbf{I}_p\right)\left(\mathbf{I}_n\otimes \mathbf{V}^\top\right)\left(\bPsi_{n\times n} \otimes \mathbf{I}_p + \mathbf{I}_n \otimes \mathbf{V}\Lambda_2\mathbf{V}^\top\right)\\
  &=\left(\mathbf{U}\otimes \mathbf{I}_p\right)\mathbf{D}\left(\mathbf{U}^\top \otimes \mathbf{I}_p\right)\left(\bPsi_{n\times n}\otimes \mathbf{V}^\top + \mathbf{I}_n \otimes \Lambda_2 \mathbf{V}^\top\right)\\ 
  &=\mathbf{I}_n \otimes \mathbf{V}^\top.  
\end{split}
\end{equation*}
\subsubsection*{A.2 Proof of Equation (\ref{eq:womegaexp})}
Note that $\mathbf{W}\bOmega=\mathbf{I}_{np}$, therefore we can write Equation (\ref{eq:wup}) as:
\[\left(\mathbf{I}_n\otimes \mathbf{V}^\top\right)\mathbf{W}\bOmega=\left(\mathbf{U}\otimes \mathbf{I}_p\right)\mathbf{D}\left(\mathbf{U}^\top \otimes \mathbf{V}^\top\right)\bOmega.\]
Multiply both sides of the equation above by $\mathbf{I}_n \otimes \mathbf{V}$:
\[\left(\mathbf{I}_n\otimes \mathbf{V}^\top\right) \mathbf{W} \bOmega\left(\mathbf{I}_n \otimes \mathbf{V}\right)=\left(\mathbf{U}\otimes \mathbf{I}_p\right)\mathbf{D}\left(\mathbf{U}^\top \otimes \mathbf{V}^\top\right)\bOmega\left(\mathbf{I}_n \otimes \mathbf{V}\right).\]
From the right-hand side, we get $\left(\mathbf{U}\otimes \mathbf{I}_p\right)\mathbf{D}\left(\mathbf{U}^\top \otimes \mathbf{I}_p\right)\left(\bPsi_{n\times n}\otimes \mathbf{I}_p+ \mathbf{I}_n \otimes \Lambda_2\right) $. On the left-hand side, remember that $\mathbf{W}\bOmega=\mathbf{I}_{np}$, so \[\left(\mathbf{I}_n\otimes\mathbf{V}^{\top}\right)\mathbf{W}\bOmega\left(\mathbf{I}_n\otimes\mathbf{V}\right)=\mathbf{I}_n\otimes\mathbf{I}_p.\]
Indeed we get Equation (\ref{eq:womegaexp}) in the main paper.
\subsubsection*{A.3 Proof of Equation (\ref{eq:W1_not1})}
In order to prove Equation (\ref{eq:W1_not1}), we first note that, from the bottom-left block of
\begin{equation*} 
  \begin{split}
    \hat{\mathbf{W}}\hat{\mathbf{\bOmega}}
    &=\begin{bmatrix}
       \hat{\mathbf{W}}_{11}    &\hat{\mathbf{W}}_{1\sm1} \\
       \hat{\mathbf{W}}_{\sm1 1} & \hat{\mathbf{W}}_{\sm1 \sm1}
     \end{bmatrix}
    \begin{bmatrix}
      \psi_{11}\mathbf{I}_p + \Lambda_2 	&\dots 	&\psi_{1n}\mathbf{I}_p \\
      \vdots 				&\ddots 	&\vdots 		\\
      \psi_{n1}\mathbf{I}_p			&\dots	&\psi_{nn}\mathbf{I}_p + \Lambda_2
    \end{bmatrix} 
    = \mathbf{I}_{n} \otimes \mathbf{I}_p
  \end{split}
\end{equation*}
we get
\begin{align*}
  \hat{\mathbf{W}}_{\sm1 1} \hat{\bOmega}_{11}+ \hat{\mathbf{W}}_{\sm1\sm1} \hat{\bOmega}_{\sm1 1} &=
  \hat{\mathbf{W}}_{\sm1 1} \left(\psi_{11}\mathbf{I}_p + \Lambda_2\right) + \hat{\mathbf{W}}_{\sm1\sm1} \left(\bpsi_{\sm1 1} \otimes \mathbf{I}_p\right)=\mathbf{0}_{n-1} \otimes \mathbf{I}_p.
\end{align*}
Thus, multiplying both sides of the last equation by $\left(\psi_{11}\mathbf{I}_p + \Lambda_2\right)^{-1}$, one has
\begin{equation*}
  \hat{\mathbf{W}}_{\sm1 1} + \hat{\mathbf{W}}_{\sm1\sm1}
  \begin{bmatrix}
    \left(\psi_{11}\mathbf{I}_p + \Lambda_2\right)^{-1} \psi_{21} \\
    \vdots \\
    \left(\psi_{11}\mathbf{I}_p + \Lambda_2\right)^{-1}\psi_{n1}
  \end{bmatrix}
  = \mathbf{0}_{n-1} \otimes \mathbf{I}_p.
\end{equation*}
\subsubsection*{A.4 Proof of Proposition 3.1}
Proposition 3.1 follows from the fact that
\begin{equation*}
  \left[\mathbf{I}_n \otimes \mathbf{V}^\top\right] \mathbf{W} \left[\mathbf{I}_n\otimes \mathbf{V} \right]=\left[\mathbf{U}\otimes \mathbf{I}_p\right]\mathbf{D}\left[\mathbf{U}^\top \otimes \mathbf{I}_p\right]=\hat{\mathbf{W}}.
\end{equation*}
Then, the $p \times p$ blocks of $\mathbf{W}$ and $\hat{\mathbf{W}}$ hold a similarity relation:
\begin{equation*}
  \hat{\mathbf{W}}_{ij}=\mathbf{V}^\top\mathbf{W}_{ij}\mathbf{V}
\end{equation*} 
and hence  $\text{tr}_p\left(\mathbf{W}\right)= \text{tr}_p\left(\hat{\mathbf{W}}\right)$. 
\subsubsection*{A.5 Proof of Proposition 3.2}
To prove Proposition 3.2, we note that
\begin{equation*}
  \hat{\mathbf{W}}_{\sm1\sm1}=\left[\mathbf{U}_{\sm1}\otimes \mathbf{I}_p \right]\mathbf{D} \left[\mathbf{U}_{\sm1}^\top\otimes \mathbf{I}_p\right]=
  \begin{bmatrix}
    u_{21}\mathbf{I}_p & \ldots &u_{2n}\mathbf{I}_p  \\
    \vdots & \ddots & \vdots \\
    u_{n1}\mathbf{I}_p & \ldots &u_{nn}\mathbf{I}_p\end{bmatrix}
    \mathbf{D}
    \begin{bmatrix}
      u_{21}\mathbf{I}_p & \ldots &u_{n1}\mathbf{I}_p  \\
      \vdots & \ddots & \vdots \\
      u_{2n}\mathbf{I}_p & \ldots &u_{nn}\mathbf{I}_p
    \end{bmatrix},
\end{equation*}

where $\mathbf{U}_{\sm1}$ $\in \mathbb{R}^{\left(n-1\right) \times n}$ is the matrix formed by the last $n-1$ rows of $\mathbf{U}$. Then, we can decompose  $\hat{\mathbf{W}}_{\sm1\sm1}$ in $\left(n-1\right)\times\left(n-1\right)$ blocks  
$[\hat{\mathbf{W}}_{\sm1\sm1}]_{\ell,k} \in \mathbb{R}^{p \times p}$, with
%
\begin{equation*}
  [\hat{\mathbf{W}}_{\sm1\sm1}]_{\ell,k} =
  \begin{bmatrix}
    \sum^{n}_{i=1}\frac{\  u_{\left(\ell\right)i}u_{k i}}{\lambda_{1i} + \lambda_{21 }}& \ldots & 0\\
    
    0 & \ldots & \sum_{i=1} ^n \frac{u_{\left(\ell\right)i}u_{k i}}{\lambda_{1i} + \lambda_{2p } }
  \end{bmatrix},\quad \ell,k \in \{2,\ldots, n\}.
\end{equation*}
%
 Note that if we partition the $ \hat{\mathbf{W}}$ into four blocks starting from any other  $ \hat{\mathbf{W}}_{hh}$ with $ h\in\left\{1,\ldots, n\right\}$ the above sums would be over $i\in\{1,\ldots,h-1,h+1,\ldots,n\}$.
 This formulation allows us to write each trace term of Equation (10) in the main paper as
%
    \begin{equation*}
       \textrm{tr}_p\left\{ \hat{\mathbf{W}}_{\sm1 k} \left(\psi_{11}\mathbf{I}_p + \bLambda_2\right)^{-1} \right\}
  =
   \begin{bmatrix}
     \textrm{tr}\left\{\hat{\mathbf{W}}_{\sm1\sm1}\right\}_{1,k} \left(\psi_{11}\mathbf{I}_p + \bLambda_2\right)^{-1} \\
     \vdots& \\
     \textrm{tr}\left\{\hat{\mathbf{W}}_{\sm1\sm1}\right\}_{\left(n-1\right),k} \left(\psi_{11}\mathbf{I}_p + \bLambda_2\right)^{-1}
   \end{bmatrix} ,\quad k\in\{1,\dots,n-1\}, 
    \end{equation*}
   

  More explicitly, 
%
%
 \begin{equation*}
 \label{eqn:trace}
   \begin{split}
      \textrm{tr}_p\left\{ \hat{\mathbf{W}}_{\sm1 k} \left(\psi_{11}\mathbf{I}_p + \bLambda_2\right)^{-1} \right\}&=
  \nonumber   \begin{bmatrix}
     \sum_{j=1}^p  \sum_{i=1}^n \frac{1}{\psi_{11}+\lambda_{2j}} \frac{u_{2i}u_{ki}}{\lambda_{1i}+\lambda_{21}} \\
     \vdots& \\
     \sum_{j=1}^p  \sum_{i=1}^n \frac{1}{\psi_{11}+\lambda_{2j}} \frac{u_{ni}u_{ki}}{\lambda_{1i}+\lambda_{2p}} 
     \end{bmatrix}= 
      \begin{bmatrix}
     \sum_{j=1}^p \frac{1}{\psi_{11}+\lambda_{2j}}  \sum_{i=1}^n \frac{u_{2i}u_{ki}}{\lambda_{1i}+\lambda_{21}} \\
     \vdots& \\
     \sum_{j=1}^p \frac{1}{\psi_{11}+\lambda_{2j}}  \sum_{i=1}^n \frac{u_{ni}u_{ki}}{\lambda_{1i}+\lambda_{2p}} 
     \end{bmatrix}\\&=
     \sum_{j=1}^p \frac{1}{\psi_{11}+\lambda_{2j}}
     \nonumber  \begin{bmatrix}
       \sum_{i=1}^n \frac{u_{2i}u_{ki}}{\lambda_{1i}+\lambda_{21}}\\
       \vdots& \\
       \sum_{i=1}^n \frac{u_{ni}u_{ki}}{\lambda_{1i}+\lambda_{2p}}
     \end{bmatrix}.
     \end{split}  
 \end{equation*}
\subsubsection*{A.6 Proof of Proposition 3.3}
Proposition 3.3 follows from the fact that 
\[\mathbf{W} =\bOmega^{-1}= \left(\mathbf{U}\otimes \mathbf{V}\right) [\bLambda_1\otimes \mathbf{I}_p + \mathbf{I}_n \otimes \bLambda_2] ^{-1}\left(\mathbf{U}^\top \otimes \mathbf{V}^\top\right),\]
and 
\[\mathbf{D} = \begin{bmatrix}
    \frac{1}{\lambda_{11}+\lambda_{21} } & \dots & 0 & \dots & 0 & \dots & 0 \\
    \vdots & \ddots & \vdots & \dots & \vdots & \dots & \vdots \\
   0 & \dots & \frac{1}{\lambda_{11}+\lambda_{2p} } & \dots & 0 & \dots & 0 \\
    \vdots & \dots & \vdots & \ddots & \vdots & \dots & \vdots \\
    0 & \dots & 0 & \dots & \frac{1}{\lambda_{1n}+\lambda_{21} } & \dots & 0\\
   \vdots & \dots & \vdots & \dots & \vdots & \ddots & \vdots \\
   0 & \dots & 0 & \dots & 0 & \dots & \frac{1}{\lambda_{1n}+\lambda_{2p} }
   \end{bmatrix},
\]
where  $\lambda_{11} \ldots \lambda_{1n}$ are the diagonal values of $\bLambda_1\in \mathbb{R}^{n\times n}$ and $\lambda_{21} \ldots \lambda_{2p}$ are the diagonal values of $\bLambda_2 \in \mathbb{R}^{p \times p}$. Then, we can write
\begin{align*}
\nonumber | \bPsi_{n\times n} \oplus \bTheta_{p\times p} | &=  |\left(\mathbf{U}\otimes \mathbf{V}\right)\mathbf{D}^{-1}\left(\mathbf{U}^\top \otimes \mathbf{V}^\top\right) |=
 |\mathbf{U}\otimes \mathbf{V}|^2|\mathbf{D}^{-1}|=
\nonumber |\mathbf{U}|^{2p}|\mathbf{V}|^{2n}\prod_{i=1}^n\prod_{j=1}^p\left(\lambda_{1i}+\lambda_{2j}\right)\\&=
\nonumber \prod_{i=1}^n\prod_{j=1}^p\left(\lambda_{1i}+\lambda_{2j}\right).
\end{align*}

\subsubsection*{A.7 Some mathematical details for Section 4.1}
Consider the $p\times n$ random matrix $\mathbf{Y}=\left(Y_{ij}\right)\;,i=1,\dots,p,\;j=1,\dots,n$.
Consider for each row vectors of $\bf{Y}$, $Y_i=\left(Y_{i1},\dots,Y_{in}\right)^\top,\;i=1,\dots,p$, the marginal distributions $F^{\left(r\right)}_{1},\dots,F^{\left(r\right)}_{j},\dots,F^{\left(r\right)}_{n}$, where the superscript $(r)$ denotes marginal distributions in row vector. Then by Sklar's theorem, for a $n-$dimensional distribution function $\Phi_{\left(\bf{0}_{n},\bPsi_{n\times n}^{-1}\right)}$, there exists copula $C^{\left(r\right)}$ such that 
\begin{equation*}
\Phi_{\left\{\bf{0}_{n},\bPsi_{n\times n}^{-1}\right\}}\left(\Phi^{-1}\left(F^{\left(r\right)}_1\left(Y_{i1}\right)\right),\dots,\Phi^{-1}\left(F^{\left(r\right)}_n\left(Y_{in}\right)\right)\right)
=C^{\left(r\right)}\left(F^{\left(r\right)}_1\left(Y_{i1}\right),\dots,F^{\left(r\right)}_n\left(Y_{in}\right)\right). 
\end{equation*}
That is, there exist functions $f^{\left(r\right)}=\left\{f^{\left(r\right)}_j\right\}_{j=1}^{n}$ such that for each row vectors of $\bf{Y}$, $Y_i=\left(Y_{i1},\dots,Y_{in}\right)^\top,\;i=1,\dots,p$, $Z^{\left(r\right)}_i\equiv f^{\left(r\right)}\left(Y_i\right)\sim\mbox{mN}\left(\boldsymbol{0}_n,\bPsi_{n\times n}\right)$, where $f^{\left(r\right)}\left(Y_i\right)=\left(f^{\left(r\right)}_1\left(Y_{i1}\right),\dots,f^{\left(r\right)}_n\left(Y_{in}\right)\right)$. Then we say $Y_i=\left(Y_{i1},\dots,Y_{in}\right)^\top$ has a nonparanormal distribution and write
\[Y_i\sim\mbox{NPN}\left(\boldsymbol{0}_{n},\bPsi_{n\times n}^{-1},f^{\left(r\right)}\right).\] According to Lemma 3.1 in \cite{lafferty2012sparse}, the dependence between $Y_{i1},\;\dots,\;Y_{in}, i=1,\dots,p$, 
can be illustrated by a Gauss-Markov Graph $G_{r}=\left\{V_{r},E_r\right\}$ corresponding to precision matrix $\bPsi_{n\times n}$. This is equivalent to have latent variable $\bf{Z}^{(r)}=f^{(c)}(\bf{Y_i})\sim\mbox{mN}\left(\boldsymbol{0}_{n},\bPsi_{n\times n}^{-1}\right),i=1,\dots,p$.\\
\newline
Similarly, for each column vector of $\bf{Y}$, $Y_j=\left(Y_{1j},\dots,Y_{pj}\right)^\top,\;j=1,\dots,n$, we consider marginal distributions $F^{\left(c\right)}_{1},\dots,F^{\left(c\right)}_{i},\dots,F^{\left(c\right)}_{n}$, where the superscript $(c)$ denotes marginal distributions in column vector. Then by Sklar's theorem, for a $p-$dimensional distribution function $\Phi_{\left(\bf{0}_{p},\bPsi_{p\times p}^{-1}\right)}$, there exists copula $C^{\left(c\right)}$ such that
\begin{equation*}
   \Phi_{\left(\bf{0}_{p},\bTheta_{p\times p}^{-1}\right)}\left(\Phi^{-1}\left(F^{\left(c\right)}_1\left(Y_{1j}\right)\right),\dots,\Phi^{-1}\left(F^{\left(c\right)}_p\left(Y_{pj}\right)\right)\right)
   =C^{\left(c\right)}\left(F^{\left(c\right)}_1\left(Y_{1j}\right),\dots,F^{\left(c\right)}_n\left(Y_{pj}\right)\right).   
\end{equation*}
That is, there exist functions $f^{\left(c\right)}=\left\{f^{\left(c\right)}_i\right\}_{i=1}^{p}$ such that for each column vector of $\bf{Y}$, $Y_j=\left(Y_{1j},\dots,Y_{pj}\right)^\top,\;j=1,\dots,n$, $Z^{\left(c\right)}_j\equiv f^{\left(c\right)}\left(Y_j\right)\sim\mbox{mN}\left(\boldsymbol{0}_p,\bTheta_{p\times p}^{-1}\right)$, where $f^{\left(c\right)}\left(Y_j\right)=\left(f^{\left(c\right)}_1\left(Y_{1j}\right),\dots,f^{\left(c\right)}_p\left(Y_{pj}\right)\right)$. Then we say $Y_j=\left(Y_{1j},\dots,Y_{pj}\right)^\top$ has a nonparanormal distribution and write
\[Y_j\sim\mbox{NPN}\left(\boldsymbol{0}_{p},\bTheta_{p\times p}^{-1},f^{\left(c\right)}\right).\] 
The dependence between $Y_{1j},\;\dots,\;Y_{pj}$ can be illustrated by a Gauss-Markov Graph $G_{c}=\left\{V_{c},E_{c}\right\}$ corresponding to precision matrix $\bTheta_{p\times p}$. This is equivalent to have latent variable $\bf{Z}^{(c)}=f^{(c)}(\bf{Y_j})\sim\mbox{mN}\left(\boldsymbol{0}_{p},\bTheta_{p\times p}^{-1}\right),j=1,\dots,n$.\\
\par Consider the Cartesian product between $G_{c}$ and $G_{r}$:
\[G_{c}\Box G_{r}=\left(V_{r}\times V_{c},\{(v_1,v_2),(v_1,v_2^{'})|v_1\in G_{c},(v_2,v_2^{'})\in E_{r} \}\bigcup\{(v_1,v_2),(v_1^{'},v_2)|v_2\in G_{r},(v_1,v_1^{'})\in E_{c} \}\right).\]
According to Theorem 4.3.5 in \cite{knauer2019algebraic} (where Cartesian product we defined here was called Box product), the mapping $V_1\times V_2\longrightarrow G_{c}\Box G_{r}$ is bimorphism. 
\par From the perspective of Gauss-Markov graph, we propose to view that after the Cartesian product of $G_{c}$ and $G_{r}$, the latent variables was mapped to a new set of latent variable $\bf{Z}$ for the total mateix $\bf{Y}$, $Z^{(c)}\times Z^{(r)}\longrightarrow$ Z. As in Gauss-Markov graph, the support of precision matrix defines the adjacency matrix of the corresponding graph, and by \cite{cvetkovic1998spectra}, Cartesian product of graphs (Reffered to as "sum" in \cite{cvetkovic1998spectra}) corresponds to the Kronecker sum of Adjacency matrices. Therefore, the Cartesian product of Gauss-Markov graphs corresponds to the Kronecker sum of precision matrices. 
\par Assume the overall graph illustrating relationships inside $\bf{Y}$ is the Cartesian product of the graph $G_{r}$ and $G_{c}$, denoted as $G_{r}\Box G_{c}$. Then the overall graph $G_{r}\Box G_{c}$ is a Gauss-Markov Graph corresponding to precision matrix $\bPsi_{n\times n}\oplus\bTheta_{p\times p}$. 
Then we can assume for each $Y_{ij}$, there exists functions $f=\left\{f_{ij}\right\}_{\{i,j\}}$ and the latent variable $Z_{ij}=f_{ij}\left(Y_{ij}\right)$ such that $Z^{(c)}\times Z^{(r)}\longrightarrow \bf{Z}$, and
\[\mbox{vec}\left(\bf{Z}\right)\equiv f\left(\mbox{vec}\left(\bf{Y}\right)\right)\sim\mbox{mN}\left(\boldsymbol{0}_{np},\bOmega^{-1}\right),\] where $\bOmega=\bPsi_{n\times n}\oplus \bTheta_{p\times p}$ is the corresponding precision matrix.

\subsection*{B The effect of regularization parameters}

\begin{figure}[h]
\vspace{-20pt}
    \begin{center}
        \includegraphics[width=\textwidth]{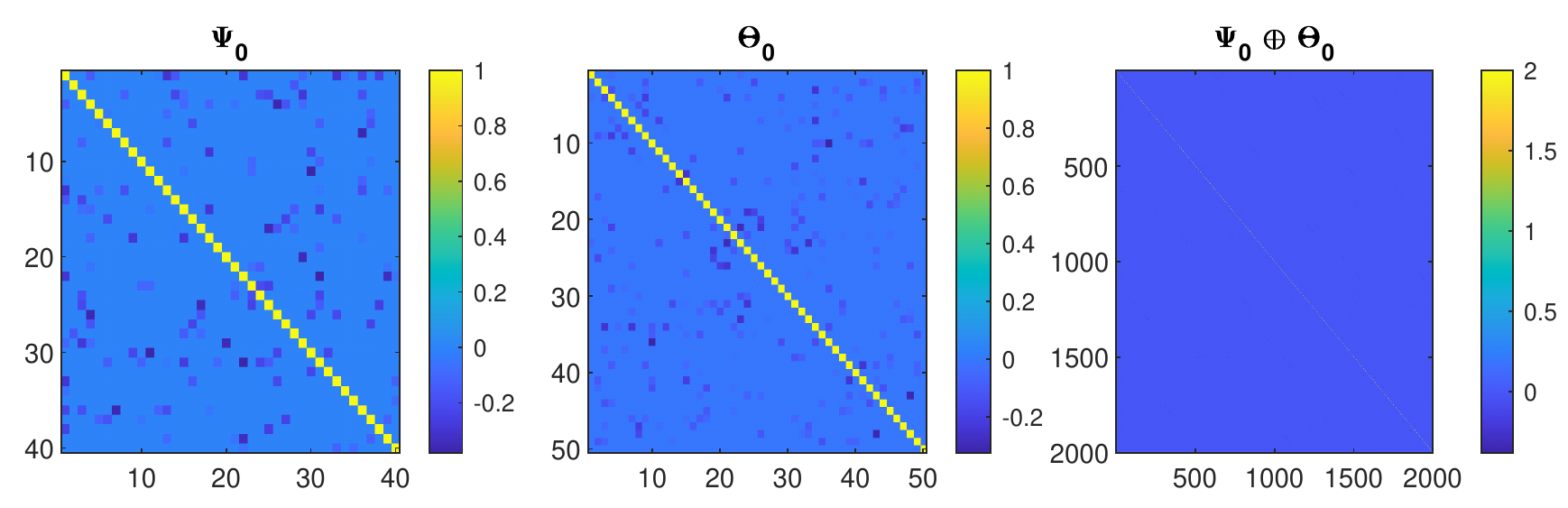}
    \caption{Precision matrix $\bPsi_0$ (left),  $\bTheta_0$ (centre) and corresponding Kronecker product matrix $\Omega_0$ (right) for our exemplar synthetic dataset. }
    
    \label{fig:synth_Psi0_Theta0_Omega0}
    \end{center}
\end{figure}

Our algorithms depend on the regularization parameters $\beta_1$ and $\beta_2$. Figure \ref{fig:synth_result_BIC} below illustrates the effect of these parameters on the performance of our algorithms. We generated two random sparse positive-definite matrices with a sparsity of 0.1 and non-zero entries normally distributed with mean 1 and variance 2. These were used as precision matrices $\bPsi_0$ and $\bTheta_0$ to create the Kronecker product matrix $\Omega_0$ as plotted in Figure \ref{fig:synth_Psi0_Theta0_Omega0}. This synthetic dataset corresponds to the experiment plotted in Figure \ref{fig:synth_result} of our paper.

\begin{figure}[h]
\vspace{-20pt}
    \begin{center}
        \includegraphics[width=\textwidth]{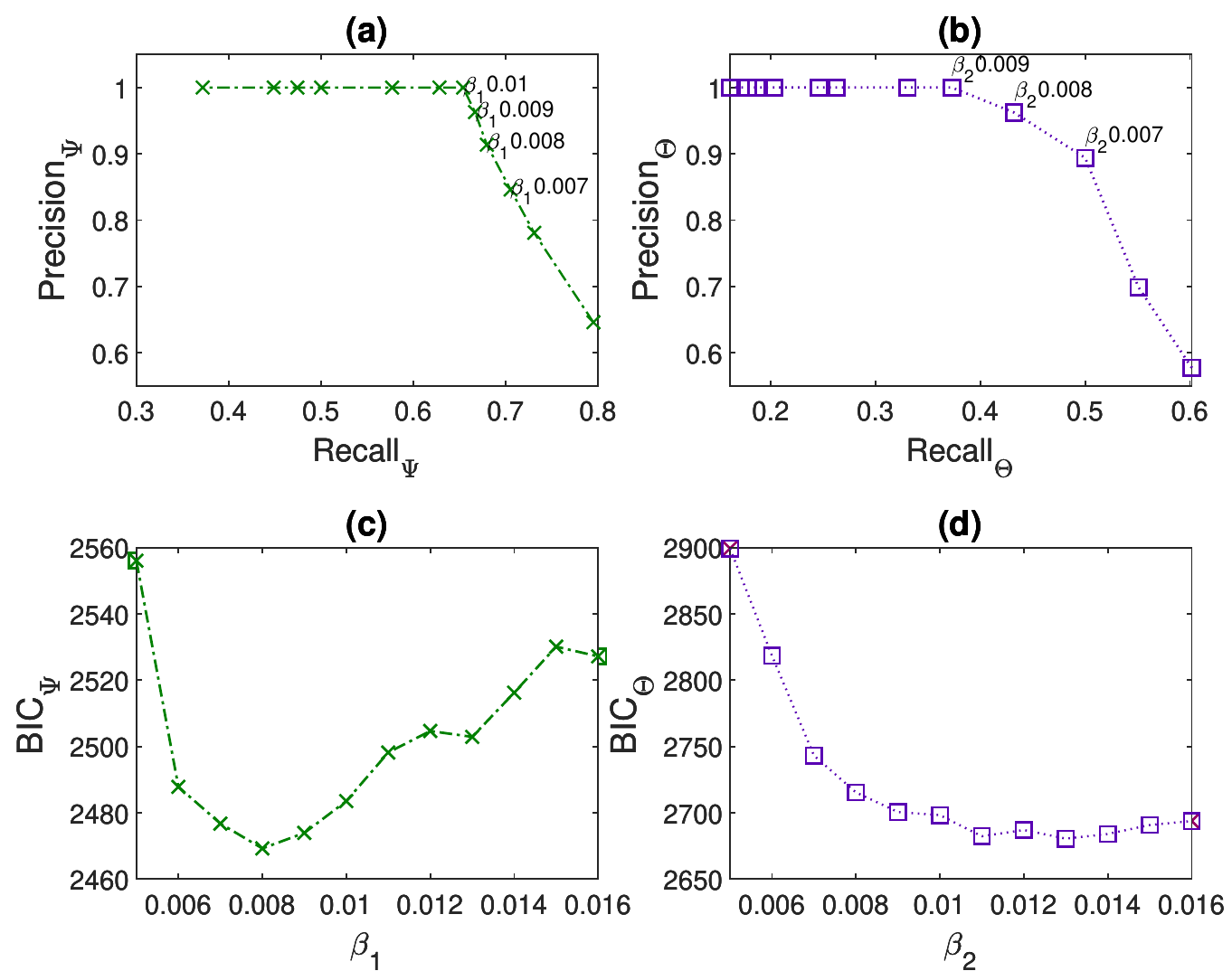}
    \caption{Synthetic network recovery results. Bayesian Information Criterion and regularization parameters.{\bf (a)} Precision-Recall of the network recovery relating to the support of $\bPsi_{n\times n}$; {\bf (b)} Precision-Recall of the network recovery relating to the support of $\bTheta_{p\times p}$;\\
    Bayesian Information Criterion and regularization parameters.
    {\bf (c)}  $\beta_1$-$BIC_{\Psi}$; {\bf (d)} $\beta_2$-$BIC_{\Theta}$;}
    
    \label{fig:synth_result_BIC}
    \end{center}
\end{figure}

In Figure \ref{fig:synth_result_BIC}(a)-(b) we show the Precision-Recall for the estimated precision matrices. In particular, subfigure (a) refers to the estimate of $\bPsi_{n\times n}$ when varying $\beta_1$, while subfigure (b) refers to the estimate of $\bTheta_{p\times p}$ when varying $\beta_2$. These curves suggest that optimal choices of $\beta_1$ lie within the interval $[0.007,0.01]$ and similarly $\beta_2$ should lie within the interval $[0.006,0.008]$. When choosing values within these intervals, one tries to strike a balance between Precision and Recall. In order to explore further the impact of the regularization parameters, we also computed the Bayesian Information Criteria ($BIC$) described in \cite{schwarz1978estimating}. In subfigures (c) and (d) we plot the BIC curves corresponding to the estimated precision matrices when varying $\beta_1$ and $\beta_2$ respectively. BIC is an heuristic criteria that helps selecting from several models. Ones with lower BIC values are generally preferred, however, a lower BIC does not necessarily indicate one model is better than another and further investigation is usually needed. The BIC curve depicted in subfigure (c) confirms the suggestion on the optimal choices for the regularization parameters obtained with the Precision-Recall plot, but the BIC curve in subfigure (d) suggest a different range for optimal regularization parameter in $[0.01,0.016]$. Therefore, when dealing with problems without known truth, although BIC can be used to help identify the interval of potential optimal regularization parameters, it is not necessarily accurate and should be used with caution. Alternative methods to find the optimal regularization parameter should be explored in the future.

\end{document}